\documentclass[a4paper, 9pt]{article}

\usepackage{geometry}
\usepackage[numbers]{natbib}
\usepackage[utf8]{inputenc} 
\usepackage[T1]{fontenc}    
\usepackage[english]{babel}
\usepackage{graphicx}
\usepackage{framed}
\usepackage[normalem]{ulem}
\usepackage{amsmath}
\usepackage{amssymb}
\usepackage{amsfonts}

\usepackage{amsthm}
\usepackage{caption}
\usepackage{subcaption}
\usepackage{float}
\usepackage{verbatim}
\usepackage{braket}
\usepackage{mathtools}
\usepackage{nicefrac}
\usepackage{indentfirst}
\usepackage{url}
\usepackage[colorlinks]{hyperref}
\hypersetup{
  allcolors = [rgb]{0.2,0.5,0.6},
  citecolor = [rgb]{0.8,0.2,0.4}
}
\usepackage{nameref}
\usepackage{authblk}
\usepackage{bbm}
\usepackage{mdframed}
\usepackage{pgf}
\usepackage{pgfplots}
\pgfplotsset{compat=1.5}
\usepackage{appendix}

\newmdtheoremenv{theorem}{Theorem}
\newmdtheoremenv{lemma}{Lemma}

\renewcommand{\cite}[1]{\citep{#1}}

\newcommand{\vl}{\underline}
\newcommand{\mc}{\mathcal}
\newcommand{\mb}{\mathbb}
\newcommand{\mr}{\mathrm}
\newcommand{\tl}{\tilde}
\newcommand{\LA}{\left\langle}
\newcommand{\RA}{\right\rangle}
\newcommand{\abs}[1]{|\!|#1|\!|}
\newcommand{\eref}[1]{Eq.~(\ref{#1})}
\newcommand{\tref}[1]{Theorem~\ref{#1}}
\newcommand{\lref}[1]{Lemma~\ref{#1}}
\newcommand{\aref}[1]{Appendix~\ref{#1}}
\newcommand{\sref}[1]{Section~\ref{#1}} 
\newcommand{\fref}[1]{Fig.~\ref{#1}}

\title{Asymptotic learning curves of kernel methods:  empirical data {\it v.s.}  Teacher-Student paradigm}
\author[a]{Stefano Spigler}
\author[a]{Mario Geiger}
\author[a]{Matthieu Wyart}
\affil[a]{Institute of Physics, \'Ecole Polytechnique F\'ed\'erale de Lausanne, 1015 Lausanne, Switzerland}

\begin{document}
\maketitle

\begin{abstract}
How many training data are needed to learn a supervised task? It is often observed that the generalization error decreases  as $n^{-\beta}$ where $n$ is the number of training examples and $\beta$  an exponent that  depends on both data and algorithm. In this work we measure  $\beta$  when applying kernel methods to real datasets. For MNIST we find $\beta\approx 0.4$ and for CIFAR10 $\beta\approx 0.1$, for both regression and classification tasks, and for Gaussian or Laplace kernels. To rationalize the existence of non-trivial exponents that can be independent of the specific kernel used, we study the Teacher-Student framework for kernels. In this scheme, a Teacher generates data according to a Gaussian random field, and a Student learns  them via kernel regression. With a simplifying assumption --- namely that the data are sampled from a regular lattice --- we derive analytically $\beta$  for translation invariant kernels, using previous results from the kriging literature.  Provided that the Student is not too sensitive to high frequencies, $\beta$ depends only on the smoothness and dimension of the training data. We confirm numerically that these predictions hold when the training points are  sampled  at random on a hypersphere. Overall, the test error is found to be controlled by the magnitude of the projection of the true function on the kernel eigenvectors whose rank is larger than $n$.
Using this idea we predict  the exponent $\beta$ from real data by 
performing kernel PCA, leading to $\beta\approx0.36$ for MNIST and $\beta\approx0.07$ for CIFAR10, in good agreement with observations. We argue that these rather large exponents are possible due to the small effective dimension of the data.
\end{abstract}

\section{Introduction}

In supervised learning machines learn from a finite collection of $n$ training data, and their generalization error is then evaluated  on unseen data drawn from the same distribution. How many data are needed to learn a task is characterized by the learning curve relating generalization error to $n$. In various cases, the generalization error decays as a power law $n^{-\beta}$, with an exponent $\beta$ that depends on both the data and the algorithm. In~\cite{hestness2017deep} $\beta$ is reported for state-of-the-art (SOTA) deep neural networks for various tasks: in for \emph{neural-machine translation}  $\beta\approx0.3$--$0.36$ (for fixed model size) or $\beta\approx0.13$ (for best-fit models at any $n$); \emph{language modeling} shows $\beta\approx0.06$--$0.09$; in \emph{speech recognition}  $\beta\approx0.3$; SOTA models for \emph{image classification} (on ImageNet) have exponents $\beta\approx0.3$--$0.5$. Currently there is no available theory of  deep learning to rationalize these observations. Recently it was shown that for a proper initialization of the weights, deep learning in the infinite-width limit \cite{jacot2018neural} converges to kernel learning. Moreover, it is nowadays part of the lore that there exist kernels whose performance is nearly comparable to deep networks~\cite{bruna2013invariant,arora2019exact}, at least for some tasks. It is thus of great interest to understand the learning curves of kernels. For regression, if  the  target function being learned is simply assumed to be Lipschitz, then the best guarantee is $\beta=\nicefrac1d$~\cite{luxburg2004distance,bach2017breaking} where $d$ is the data dimension. Thus for large $d$, $\beta$ is very small: learning is completely inefficient, a phenomenon referred to as the \emph{curse of dimensionality}.
As a result, various works on kernel regression  make the much stronger assumption that the training points are sampled from a target function that belongs to the \emph{reproducing kernel Hilbert space} (RKHS) of the kernel (for Gaussian r \cite{smola1998connection}). With this assumption $\beta$ does not depend on $d$ (for instance in ~\cite{rudi2017generalization} $\beta=1/2$  is guaranteed). Yet,  RKHS is a very strong assumption which requires the smoothness of the target function to increase with $d$  \cite{bach2017breaking} (for Gaussian random fields see \aref{app:rkhssmooth}), which may not be realistic in large dimensions. 

In \sref{sec:genrealdata} we compute $\beta$ empirically for kernel methods applied on MNIST and CIFAR10 datasets. We find
$\beta_\mr{MNIST}\approx0.4$ and $\beta_\mr{CIFAR10}\approx0.1$ respectively. Quite remarkably, we observe essentially the same  exponents for regression and classification tasks, using either a Gaussian  or a Laplace kernel.  
Thus the exponents are not as small as $\nicefrac1d$ ($d=784$ for MNIST, $d=3072$ for CIFAR10), but neither are they $\nicefrac12$ as one would expect under the RKHS assumption.  These facts call for frameworks in which assumptions on the smoothness of the data can be intermediary between Lipschitz and RKHS. Here we study such a framework for regression, in which the target function is assumed to be a Gaussian random field of zero mean with translation-invariant isotropic covariance $K_T(\vl x)$. The data can equivalently be thought as being  synthesized by a ``Teacher'' kernel $K_T(\vl x)$.  Learning is performed with a ``Student'' kernel $K_S(\vl x)$ that minimizes the mean-square error. In general $K_T(\vl x)\neq K_S(\vl x)$. In this set-up learning is very similar to a technique  referred to as \emph{kriging}, or Gaussian process regression, originally developed in the geostatistics community~\cite{matheron1963principles,stein2012interpolation}.

To quantify learning, in \sref{sec:gentsproblem} we first perform numerical experiments  for data points distributed uniformly at random on a hypersphere of varying dimension $d$, focusing on a Laplace kernel for the Student, and considering a Laplace or Gaussian kernel for the Teacher. We observe that in both cases $\beta(d)$ is a decreasing function.
In \sref{sec:analyticasymptotics}, to derive $\beta(d)$  we consider the simplified situation where the Gaussian random field is sampled at training points lying on a regular lattice. Building on the kriging literature ~\cite{stein2012interpolation}, we show that $\beta$ is controlled by the high-frequency scaling of both the Teacher and Student kernels: assuming that the Fourier transforms of the kernels decay as $\tl K_T(\vl w) = c_T \abs{\vl w}^{-\alpha_T} + o\left(\abs{\vl w}^{-\alpha_T}\right)$ and $\tl K_S(\vl w) = c_S \abs{\vl w}^{-\alpha_S} + o\left(\abs{\vl w}^{-\alpha_S}\right)$, we obtain
\begin{equation}
\beta = \frac1d \min(\alpha_T-d, 2\alpha_S).\label{eq:result}
\end{equation}
Importantly (i) \eref{eq:result} leads to a prediction for $\beta(d)$ that accurately matches our numerical study for random training data points, leading to the conjecture that \eref{eq:result} holds in that case as well. We offer the following interpretation: ultimately, kernel methods are performing a local interpolation whose quality depends on the distance $\delta(n)$ between adjacent data points. $\delta(n)$ is asymptotically similar for random data or data sitting on a lattice. (ii)  If the kernel $K_S$ is not too  sensitive to  high-frequencies, then learning is optimal as far as scaling is concerned and $\beta=(\alpha_T-d)/d$. We will argue that the smoothness index $s\equiv [(\alpha_T-d)/2]$ characterizes the number of derivatives  of the target function that are continuous. We thus recover the curse of dimensionality: $s$ needs to be of order $d$ to have non-vanishing $\beta$ in large dimensions. 

We show that in some regime,   the test error for Gaussian data is controlled by an exponent $a$ describing how the coefficients of the true function in the eigenbasis of the kernel decay with rank. We estimate $a$ by the kernel principal component analysis (kernel PCA) based on diagonalizing the Gram matrix. This measure yields a prediction for the learning curve exponent $\beta$ that matches the numerical fit, with $\beta_\mr{MNIST}\approx0.36$ and $\beta_\mr{CIFAR10}\approx0.07$. 
We show in \aref{app:approxresult} using the recent  formalism of \cite{bordelon2020spectrum}, which does not assume Gaussianity but makes more technical assumptions, that the result of our theorem Eq.\ref{eq:result} is recovered, supporting further its validity  for real data.

Finally, we discuss the following apparent paradox: $\beta$ is significant for MNIST and CIFAR10, for which $d$ is a priori very large,
leading to a smoothness value $s$ in the hundreds in both cases, which appears unrealistic. In \sref{sec:randomandrealdatascaling} The paradox is resolved by considering that real datasets actually live on lower-dimensional manifolds.  As far as kernel learning is concerned, our findings support that the correct definition of dimension  should be based on  how the nearest-neighbors distance $\delta(n)$ scales with $n$: $\delta(n)\sim n^{-\nicefrac1{d_\mr{eff}}}$. Direct measurements of $\delta(n)$ support that MNIST and CIFAR10 live on manifolds of lower dimensions
$d^\mr{eff}_\mr{MNIST}\approx15$ and $d^\mr{eff}_\mr{CIFAR10}\approx35$. 

\section{Related works}

Part of the literature has investigated the problem of kernel regression from a different point of view, namely the \emph{optimal worst-case performance} (see for instance \cite{fischer2017sobolev,caponnetto2007optimal,pillaud2018statistical}). The target function is not assumed to be generated by a Gaussian random field, but its regularity is controlled using a \emph{source condition} that constrains the decay of its coefficients in the eigenbasis of the kernel. For uniform data distributions and isotropic kernels this is similar to controlling how the Fourier transform of the target function decays at high frequency. What we study in the present work is, on the contrary, the \emph{typical performance}. Indeed, it turns out that both the worst-case and the typical learning curve decay as power laws, and the latter decays faster.

The Teacher-Student framework for kernel regression was previously introduced in \cite{sollich1999learning,sollich2002gaussian}, where a formula for the learning curve was derived based on a few uncontrolled approximations. It is easy to show that their results match the predictions of our theorem, although in \cite{sollich2002gaussian} the case where the performance is limited by the Student ($\alpha_T-d>2\alpha_S$) is ignored. More recently, \cite{bordelon2020spectrum} generalized this approach using similar approximations and extended it to kernel regression applied to any target function (or ensemble thereof). Kernel PCA on MNIST was used to support that these approximations hold well on real data. An asymptotic scaling relation between $\beta$ and $a$ was obtained, again the presence of other regimes was not noted. 
By contrast we perform an exact calculations for the asymptotic behavior of Gaussian data on a lattice. 
In \aref{app:approxresult} we show that the two approaches are consistent and lead to the same  asymptotic predictions for $\beta$. 

Our set-up of Teacher-Student learning with kernels is also referred to as \emph{kriging}, or Gaussian process regression, and it was originally developed in the geostatistics community~\cite{matheron1963principles}.
In \sref{sec:analyticasymptotics} we present our theorem, that allows one to know the rate at which the test error decreases as we increase the number of training points, assumed to lie on a high-dimensional regular lattice. Similar results have been previously derived in the kriging literature~\cite{stein2012interpolation} when sampling occurs on the regular lattice with the exception of the origin, where the inference is made. Here we propose an alternative derivation that some readers might find simpler. We also study a slightly different problem: instead of computing the test error when the inference is carried on at the origin, we compute the average error for a test point that lie at an arbitrary point, sampled uniformly at random and not necessarily on the lattice. Then, in what follows we show, via extensive numerical simulations, that such predictions are accurate even when the training points do not lie on a regular lattice, but are taken at random on a hypersphere. An exact proof of our result in such a general setting is difficult and cannot be found even in the kriging literature. To our knowledge the results that get closer to the point are those discussed in~\cite{stein1999predicting}, where the author studies one-dimensional processes where the training data are not necessarily evenly spaced.

In this work the effective dimension of the data plays an import role, as it controls how the distance between nearest neighbors scales with the dataset size. Of course, there exists a vast literature ~\cite{grassberger1983measuring,costa2004learning,hein2005intrinsic,levina2005maximum,rozza2012novel,facco2017estimating,allegra2019clustering} devoted to the study of effective dimensions, where other definitions are analyzed.
The effective dimensions that we find are compatible with those obtained with more refined methods.

\section{Learning curve for kernel methods applied to real data}\label{sec:genrealdata}

In what follows we apply kernel methods to the MNIST and CIFAR10 datasets, each consisting of a set of images $(\vl x_\mu)_{\mu=1}^n$. We simplify the problem by considering only two classes whose  label $Z(\vl x_\mu)=\pm1$ correspond to odd and even numbers for MNIST, and to  two groups of 5 classes in  CIFAR10. The goal is to infer the value of the label $\hat Z_S(\vl x)$ of an image $\vl x$ that does not belong to the dataset. The $S$ subscript  reminds us that  inference is performed using a positive definite kernel $K_S$. 
We perform  inference in both a \emph{regression} and a \emph{classification} setting. The following algorithms and associated results can be found in~\cite{scholkopf2001learning}.

\paragraph{Regression.} Learning corresponds to  minimizing a mean-square error:
\begin{equation}
\min \sum_{\mu=1}^n \left[\hat Z_S(\vl x_\mu) - Z(\vl x_\mu)\right]^2.\label{eq:mse}
\end{equation}
For algorithms seeking solutions of the form $\hat Z_S(\vl x) = \sum_\mu a_\mu K_S(\vl x_\mu, \vl x) \equiv \vl a\cdot\vl k_S(\vl x)$ by minimizing the man-square loss  over the vector $\vl a$, one obtains:
\begin{equation}
\hat Z_S(\vl x) = \vl k_S(\vl x) \cdot \mb K_S^{-1} \vl Z,\label{eq:regrpred}
\end{equation}
where the vector $\vl Z$ contains all the labels in the training set, $\vl Z \equiv (Z(\vl x_\mu))_{\mu=1}^n$, and $\mb K_{S,\mu\nu} \equiv K_S(\vl x_\mu, \vl x_\nu)$ is the Gram matrix. The Gram matrix is always invertible if the kernel $K_S$ is positive definite. The generalization error is then evaluated as the expected mean-square error on unseen data, estimated by averaging over a test set composed of $n_\mr{test}$ unseen data points:
\begin{equation}
\mr{MSE} = \frac1{n_\mr{test}}\sum_{\mu=1}^{n_\mr{test}} \left[\hat Z_S(\vl x_\mu) - Z(\vl x_\mu)\right]^2.
\end{equation}

\paragraph{Classification.} We perform kernel classification via the algorithm \emph{soft-margin SVM}. The details can be found in \aref{app:smsvms}. After learning from the training data with a student kernel $K_S$, performance is evaluated via the generalization error. It is estimated as the fraction of correctly predicted labels for data points belonging to a test set with $n_\mr{test}$ elements.

\begin{figure}[ht!]
\centering
\includegraphics[width=\textwidth]{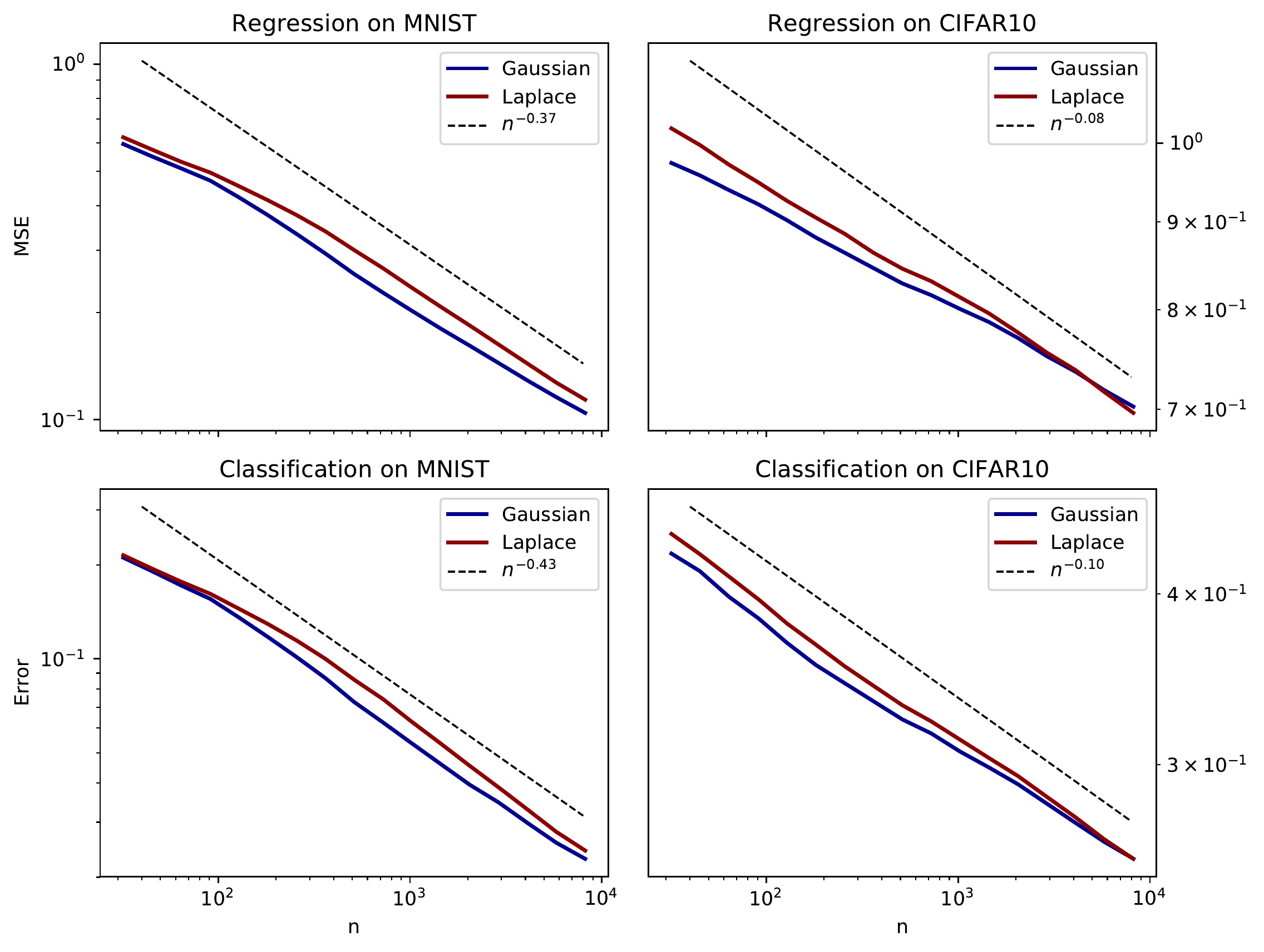}
\caption{Learning curves for regression on MNIST and CIFAR10 \emph{(\emph{top row})}; and for classification on MNIST and CIFAR10 \emph{(\emph{bottom row})}. Curves are averaged over $400$ runs. A power law is plotted to estimate the asymptotic behavior at large $n$: the exponent is fitted on the last decade on the average of the two curves, since it does not seem to depend significantly on the specific kernel or on the task. In each setting we use both a Gaussian kernel $K(\vl x) \propto \exp(-\abs{\vl x}^2/(2\sigma^2))$ and a Laplace one $K(\vl x) \propto \exp(-\abs{\vl x} / \sigma)$, with $\sigma=1000$.\label{fig:realdata}}
\end{figure}

In \fref{fig:realdata} we present the learning curves for (binary) MNIST and CIFAR10, for regression and classification. Learning is performed both with a Gaussian kernel $K(\vl x) \propto \exp(-\abs{\vl x}^2/(2\sigma^2))$ and a Laplace one  $K(\vl x) \propto \exp(-\abs{\vl x} / \sigma)$. Remarkably, the power laws in the two tasks are essentially identical (although the estimated exponent appears to be slightly larger, in absolute value, for classification). Moreover, the two kernels display a very similar behavior, compatible with the same exponent: about $-0.4$ for MNIST and $-0.1$ for CIFAR10. The presented data are for $\sigma=1000$; in \aref{app:diffsigmas} we show that the same behaviour is observed for different values.

\section{Generalization scaling in kernel Teacher-Student problems}\label{sec:gentsproblem}

We study $\beta$ in a simplified setting where the data is assumed to follow a Gaussian distribution with  known covariance. It falls into the class of teacher-Student problems, which are characterized by a machine (the Teacher) that generates the data, and another machine (the Student) that tries to learn from them. The Teacher-Student paradigm has been broadly used to study supervised learning ~\cite{saad1995line,monasson1995weight,sollich1999learning,sollich2002gaussian,opper2001advanced,engel2001statistical,barbier2017phase,gabrie2018entropy,aubin2018committee,franz2018jamming}. He we restrict our attention to kernel methods: we assume that a target function is distributed according to a Gaussian random field $Z\sim\mc N(0,K_T)$ --- the Teacher --- characterized by a translation-invariant isotropic covariance function $K_T(\vl x, \vl x^\prime) = K_T(\abs{\vl x-\vl x^\prime})$, and that the training dataset consists the finite set of $n$ observations $\vl Z = (Z(\vl x_\mu))_{\mu=1}^n$. This is equivalent to saying that the vector of training points follows a centered Gaussian distribution with a covariance matrix that depends on $K_T$ and on the location of the points $(\vl x_\mu)_{\mu=1}^n$:
\begin{equation}
\vl Z \sim \mc N\left(\vl 0, \mb K_T\right), \quad \mr{where} \quad \mb K_T = (K_T(\vl x_\mu, \vl x_\nu))_{\mu,\nu=1}^n.
\end{equation}
Once the Teacher has generated the dataset, the rest follows as in the kernel regression described in the previous section. We use another translation-invariant isotropic kernel $K_S(\vl x, \vl x^\prime)$ --- the Student --- to infer the value of the field at another point, $\hat Z_S(\vl x)$, with a regression task, i.e. minimizing the mean-square error in \eref{eq:mse}. The solution is therefore given again by \eref{eq:regrpred}.

\begin{figure}[b!]
\centering
\includegraphics[width=\textwidth]{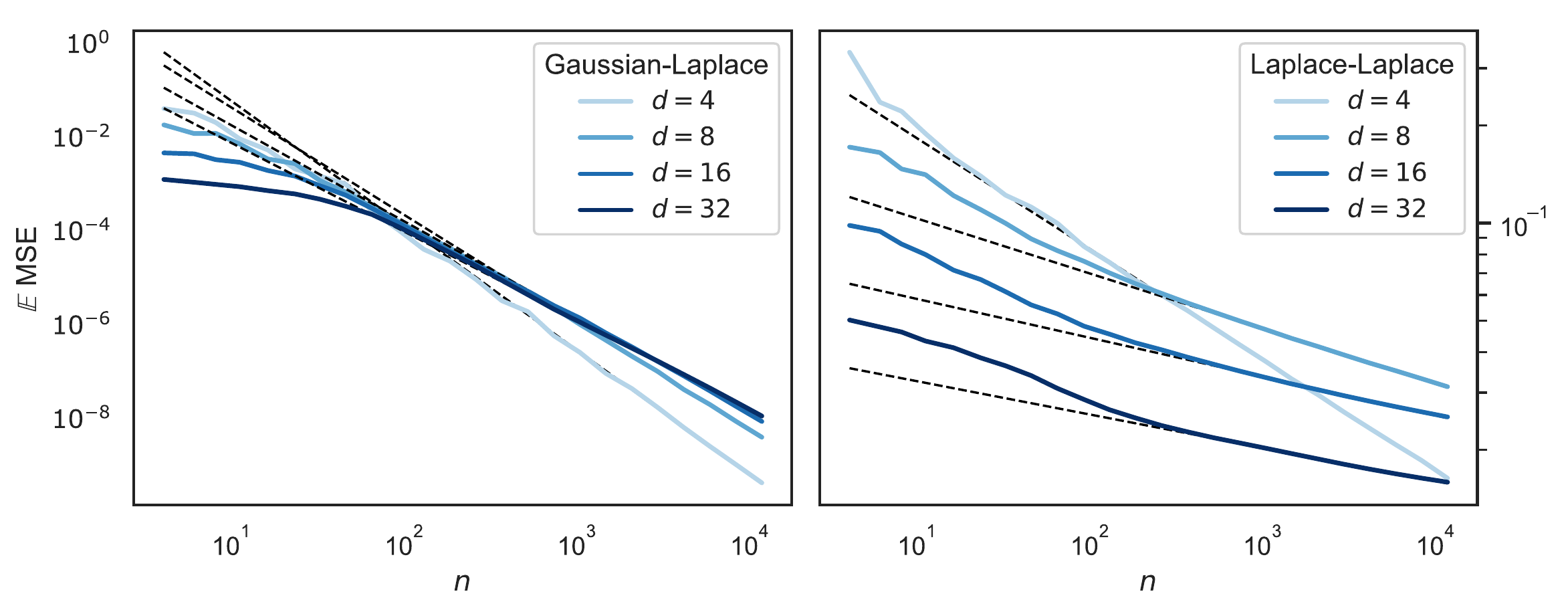}
\includegraphics[width=\textwidth]{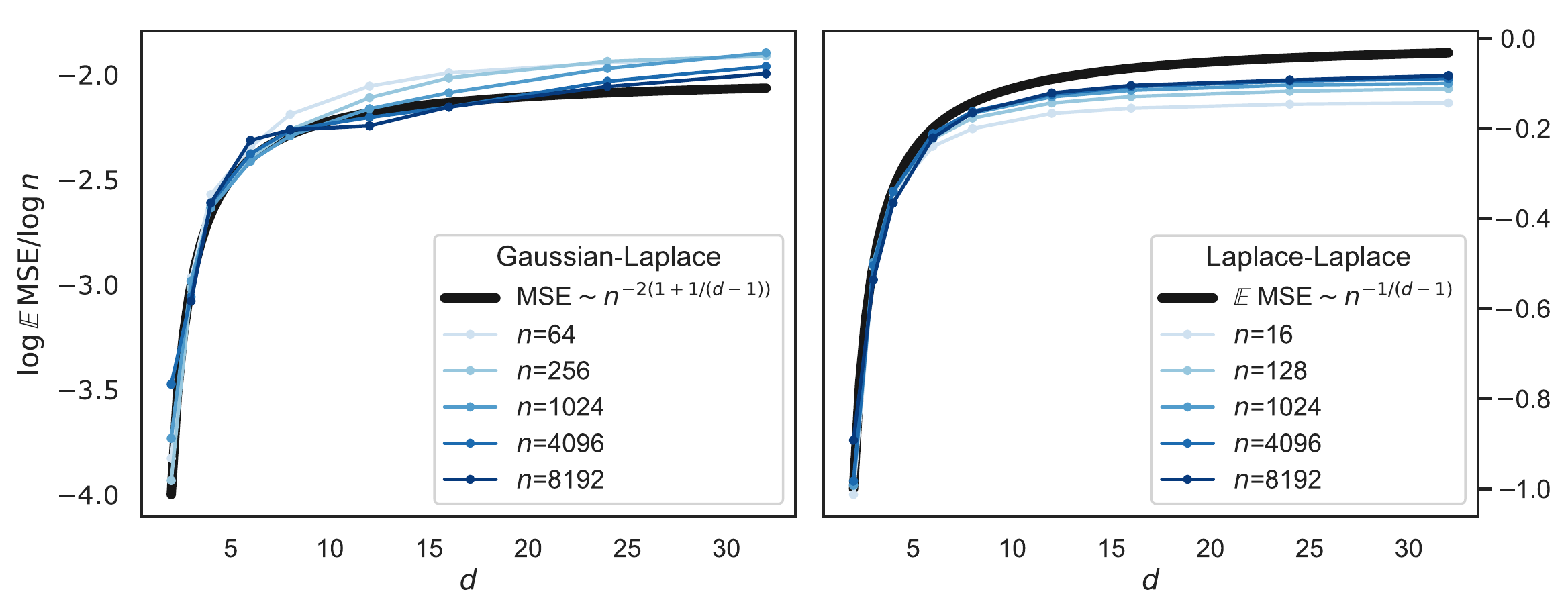}
\caption{Results for the Teacher-Student kernel regression problem, where the Student is always a Laplace kernel. Data points are sampled uniformly at random on a $d$-dimensional hypersphere. (\emph{Top row}) Mean-square error versus the size of the training dataset, for Gaussian and Laplace Teachers and for multiple spatial dimensions. Dotted lines are the fitted power laws --- we fit starting from $n=700$. (\emph{Bottom row}) Fitted exponent $-\beta=\log\mb E\,\mr{MSE} / \log n$ against the spatial dimension, for several dataset sizes. We fit from $n=0$ to a varying $n$ (written in the legends). The thick black lines are the theoretical predictions.\label{fig:kernel-exponent}}
\end{figure}

\fref{fig:kernel-exponent} \emph{(a-b)}  shows the mean-square error obtained numerically. In the examples the Student is always taken to be a Laplace kernel, and the Teacher is either a Laplace kernel or a Gaussian kernel. The points $(\vl x_\mu)_{\mu=1}^n$ are taken uniformly at random on the unit $d$-dimensional hypersphere for several dimensions $d$ and for several dataset sizes $n$. We  take $\sigma_S=\sigma_T=d$ as we observed that with this choice smaller datasets were enough to approach a limiting curve --- in \aref{app:othersigmas} we show the plots for the case $\sigma_S=\sigma_T=10$, which appears to converge to the same limit curve with increasing $n$, but at a smaller pace. The figure shows that when $n$ is large enough, the mean-square error behaves as a power law (dashed lines) with an exponent that depends on the spatial dimension of the data, as well as on the kernels. The fitted exponents are plotted in \fref{fig:kernel-exponent} \emph{(c-d)} as a function of the spatial dimension $d$ for different dataset sizes $n$. In the next section we will discuss the theoretical prediction, that in the figure is plotted a thick black line. The figure shows that as the dataset gets bigger, the asymptotic exponent tends to our prediction. In \aref{app:gaussianstudent} we present the learning curves of Gaussian Students with both a Laplace and a Gaussian kernel. When both kernels are Gaussian the test error decays exponentially fast, a result that matches our theoretical prediction. In \aref{app:matern} we also provide further numerical results for the case where the Teacher kernel is a Mat\'ern kernel (as defined therein).

\section{Analytic asymptotics for the kernel Teacher-Student problem on a lattice}\label{sec:analyticasymptotics}

In this section we compute analytically the exponent that describe the asymptotic decay of the generalization error when the number $n$ of training data increases. In order to derive the result we assume that both the Teacher Gaussian random field lives on a bounded hypercube, $\vl x\in\mc V\equiv[0,L]^d$, where $L$ is a constant and $d$ is the spatial dimension. The fields and the kernels can then be thought of as $L$-periodic along each dimension. Furthermore, to make the problem tractable we assume that the points $(\vl x_\mu)_{\mu=1}^n$ live on a \emph{regular lattice}, covering all the hypercube $\mc V$. Therefore, the linear spacing between neighboring points is $\delta = L n^{-\nicefrac1d}$. This is  a different setting than the one used in the numerical simulations presented in the previous section for which the data distribution if Gaussian,
showing that our results below are robust to such differences.

Generalization error is then evaluated via the typical mean-square error
\begin{equation}
\mb E\,\mr{MSE} = \mb E\left[Z(\vl x) - \hat Z_S(\vl x)\right]^2,
\end{equation}
where the expectation is taken over both the Teacher process and the point $\vl x$ at which we estimate the field, assumed to be uniformly distributed in the hypercube $\mc V$. In \aref{app:maintheorem} we prove the following:\vspace{0.5em}

\begin{theorem}\label{th:theorem}
Let $\tl K_T(\vl w) = c_T \abs{\vl w}^{-\alpha_T} + o\left(\abs{\vl w}^{-\alpha_T}\right)$ and $\tl K_S(\vl w) = c_S \abs{\vl w}^{-\alpha_S} + o\left(\abs{\vl w}^{-\alpha_S}\right)$ as $\abs{\vl w}\to\infty$, where $\tl K_T(\vl w)$ and $\tl K_S(\vl w)$ are the Fourier transforms of the kernels $K_T(\vl x)$, $K_S(\vl x)$ respectively, assumed to be positive definite. We assume that $\tl K_T(\vl w)$ and $\tl K_S(\vl w)$ have a finite limit as $\abs{\vl w}\to0$ and that $K(\vl 0)<\infty$. Then,
\begin{equation}
\mb E\,\mr{MSE} = n^{-\beta} + o\left(n^{-\beta}\right) \quad \mr{with} \quad \beta=\frac1d\min(\alpha_T-d, 2\alpha_S).\label{eq:maintheorem}
\end{equation}
Moreover, in the case of a Gaussian kernel the result holds valid if we take the corresponding exponent to be $\alpha=\infty$.
\end{theorem}

Apart from the specific value of the exponent in \eref{eq:maintheorem}, \tref{th:theorem} implies that if the Student kernel decays fast enough in the frequency domain, then $\beta$ depends only on the data through the behaviour of the Teacher kernel at high frequencies.  One then recovers $\beta = (\alpha_T-d)/d$, also found for the \emph{Bayes-optimal} setting where the Student is identical to the Teacher.

Consider  the predictions of \tref{th:theorem} in the cases presented in \fref{fig:kernel-exponent} \emph{(a-b)} of Gaussian and Laplace kernels. If both kernels are Laplace kernels then $\alpha_T=\alpha_S=d+1$ and $\mb E\,\mr{MSE} \sim n^{-\nicefrac1d}$, which scales very slowly with the dataset size in large dimensions. 
If the Teacher is a Gaussian kernel ($\alpha_T=\infty$) and the Student is a Laplace kernel then $\beta=2(1 + 1/d)$, leading to $\beta\rightarrow 2$ as $d\rightarrow 
\infty$. In \fref{fig:kernel-exponent} \emph{(c-d)} we compare these predictions with the exponents extracted from \fref{fig:kernel-exponent} \emph{(a-b)}. We plot $\log\mb E\,\mr{MSE} / \log n\equiv -\beta$, against the dimension $d$ of the data, varying the dataset size $n$. The exponents extracted numerically tend to our analytical predictions when $n$ is large enough.

Notice that, although the theory and the experiments do not assume the same distribution for the sampling points $(\vl x_\mu)_{\mu=1}^n$, this does not seem to yield any difference in the asymptotic behavior of the generalization error, leading to the conjecture that our predictions are exact even when the training set is random, and does not correspond to a lattice. The conjecture can be proven in one dimension following results of the kriging literature ~\cite{stein1999predicting}, but generalization to higher $d$ is a much harder problem. Intuitively, for kernel learning performs an expansion, whose quality is governed by the target function smoothness and the typical distance $\delta_{\min{}}$ between a point and its nearest neighbors in the training set. Both for random points  or on a lattice, one has $\delta_{\min{}} \sim n^{-\nicefrac1d}$ when $n$ is large enough, thus both situations lead to the same $\beta$. This is shown in \fref{fig:scaling_delta_min} (\emph{left}).


\tref{th:theorem} underlines that kernel methods are subjected to the curse of dimensionality. Indeed for appropriate students, one obtains $\beta=(\alpha_T-d)/d$. Let us define the smoothness index $s\equiv [(\alpha_T-d)/2]=\beta d/2$, which must be ${\cal O}(d)$ to avoid $\beta\rightarrow 0$ for large $d$. The two Lemmas below, derived in Appendix, indicate that the target function is $s$ time differentiable (in a mean-square sense). Thus learning with kernels in very large dimension can only occur if the target function  is ${\cal O}(d)$ times differentiable, a condition that appears very restrictive in large $d$.


\vspace{0.5em}
\begin{lemma}\label{lm:lemma1}
Let $K(\vl x, \vl x^\prime)$ be a translation-invariant isotropic kernel such that $\tl K(\vl w) = c \abs{\vl w}^{-\alpha} + o\left(\abs{\vl w}^{-\alpha}\right)$ as $\abs{w}\to\infty$ and $\abs{\vl w}^d \tl K(\vl w)\to0$ as $\abs{w}\to0$. If $\alpha > d+n$ for some $n\in\mb Z^+$, then $K(\vl x) \in C^n$, that is, it is at least $n$-times differentiable. (Proof in \aref{app:prooflemmas}).
\end{lemma}

\vspace{0.5em}
\begin{lemma}\label{lm:lemma2}
Let $Z\sim\mc N(0,K)$ be a $d$-dimensional Gaussian random field, with $K\in C^{2n}$ being a $2n$-times differentiable kernel. Then $Z$ is $n$-times \emph{mean-square} differentiable in the sense that
\begin{itemize}
\item derivatives of $Z(\vl x)$ are a Gaussian random fields;
\item $\mb E \partial_{x_1}^{n_1}\cdots\partial_{x_d}^{n_d} Z(\vl x) = 0$;
\item $\mb E \partial_{x_1}^{n_1}\cdots\partial_{x_d}^{n_d} Z(\vl x) \cdot \partial_{x_1}^{n_1^\prime}\cdots\partial_{x_d}^{n_d^\prime} Z(\vl x^\prime) = \partial_{x_1}^{n_1+n_1^\prime}\cdots\partial_{x_d}^{n_d+n_d^\prime} K(\vl x - \vl x^\prime) < \infty$ if the derivatives of $K$ exist.
\end{itemize}
In particular, $\mb E \partial_{x_i}^{m} Z(\vl x) \cdot \partial_{x_i}^{m} Z(\vl x^\prime) = \partial_{x_i}^{2m}K(\vl x-\vl x^\prime) < \infty\ \forall m\leq n$. (Proof in \aref{app:prooflemmas}).
\end{lemma}

{\bf Interpretation of} \tref{th:theorem}: When the student does not limit  performance, i.e. when $2\alpha_S > \alpha_T-d$ and $\beta = \frac{\alpha_T-d}{d}$, we can interpret the result as follows. An isotropic Student kernel corresponds to a Gaussian prior on the Fourier coefficients of the target function being learned. The student puts large (low) power at low (high) frequencies, and it can then reconstruct a number of the order of $n$ largest Fourier coefficients, which corresponds to frequencies $\vl w$ of  norm  $\abs{\vl w} \leq \nicefrac1\delta\sim n^{1/d}$. Fourier coefficients $\tilde{Z}(\vl w)$ at higher frequencies cannot be learned, and the mean square error is then simply of order of the sum of the squares of these coefficients:
\begin{equation}
    \mb E\,\mr{MSE} \sim \sum_{\abs{\vl w} \geq  n^{1/d}} |\tilde{Z}(\vl w)|^2\sim \sum_{\abs{\vl w} \geq  n^{1/d}}\abs{\vl w}^{-\alpha_T}\sim n^{\frac{d -\alpha_T}d}\sim n^{-\beta}. \label{eq:interpretfreq}
\end{equation}

\section{Learning curve exponent of real data}\label{sec:lcexponent}

 \eref{eq:interpretfreq} is not readily applicable to real data which are neither Gaussian nor uniformly distributed. However it supports the following broader result: kernel methods can predict well of order $n$ first coefficients of the true function in the eigenbasis of the kernel,
 but not the following ones. 
For any student kernel $K_S$, let  $\lambda_1\geq\lambda_\rho\geq\dots$ be its eigenvalues (positive and real, because of symmetry and positive definiteness) and $\phi_\rho(\vl x)$ the associated eigenfunctions:
\begin{equation}
    \int \mr d^d\vl y\,p(\vl y) K_S(\vl x - \vl y) \phi_\rho(\vl y) = \lambda_\rho \phi_\rho(\vl x),\label{eq:eigenker}
\end{equation}
where $p(\vl x)$ is the density of the data points. Then the kernel can be decomposed in its eigenmodes:
\begin{equation}
    K_S(\vl x - \vl y) = \sum_{\rho\geq1} \lambda_\rho \phi_\rho(\vl x) \phi_\rho(\vl y). \label{eq:eigendecomp}
\end{equation}
The eigenfunctions of $K$ make a complete basis, and we can write any function $Z(\vl x)$ as
\begin{equation}
    Z(\vl x) = \sum_\rho q_\rho \phi_\rho(\vl x). \label{eq:fielddecomp}
\end{equation}
The generalization of our result then simply reads:
\begin{equation}
    \mb E\,\mr{MSE} \sim \sum_{\rho \geq n} q_\rho^2. \label{eq:interpretmodes}
\end{equation}
Assuming a power-law behavior  $q_\rho^2\sim \rho^{-a}$ then leads to $\beta=a-1$.

To extract the exponent $a$ and test this prediction for real data, we first approximate the eigenvalue equation \eref{eq:eigenker} for the Student kernel with the diagonalization of its finite-dimensional Gram matrix $\mb K_S $ computed on a large dataset of size $\tl n$:
\begin{equation}
    \mb K_S \vl\phi_\rho \sim \lambda_\rho \vl\phi_\rho,
\end{equation}
where now we have $\tl n$ eigenvalues $\lambda_1\geq\cdots\geq\lambda_{\tl n}$ and the eigenvectors $\vl\phi_\rho$ are $\tl n$-dimensional eigenvectors. Computing this diagonalization for a given training set is referred to (uncentered)  \emph{kernel PCA}. This procedure is a discretized version of \eref{eq:eigendecomp} and yields only an approximation to the largest $\tl n$ eigenvalues of the kernel, that are exactly recovered as $\tl n\to\infty$ (in \fref{fig:distlambda} in \aref{app:otherplots} we show that the eigenvalues of the Gram matrix converge when $\tl{n}$ increases, and that their density displays the power-law behavior that one can extract from the kernel operator with a uniform distribution $p(\vl x)$). The coefficient $q_\rho$ is then estimated  by the scalar products $\left(\vl Z \cdot \vl\phi_\rho\right)$, where $\vl Z = (Z(\vl x_1), \cdots, Z(\vl x_{\tl n}))$ is the vector of the target function's values on the train set.

\begin{figure}[tbh!]
\centering
\includegraphics[width=0.5\textwidth]{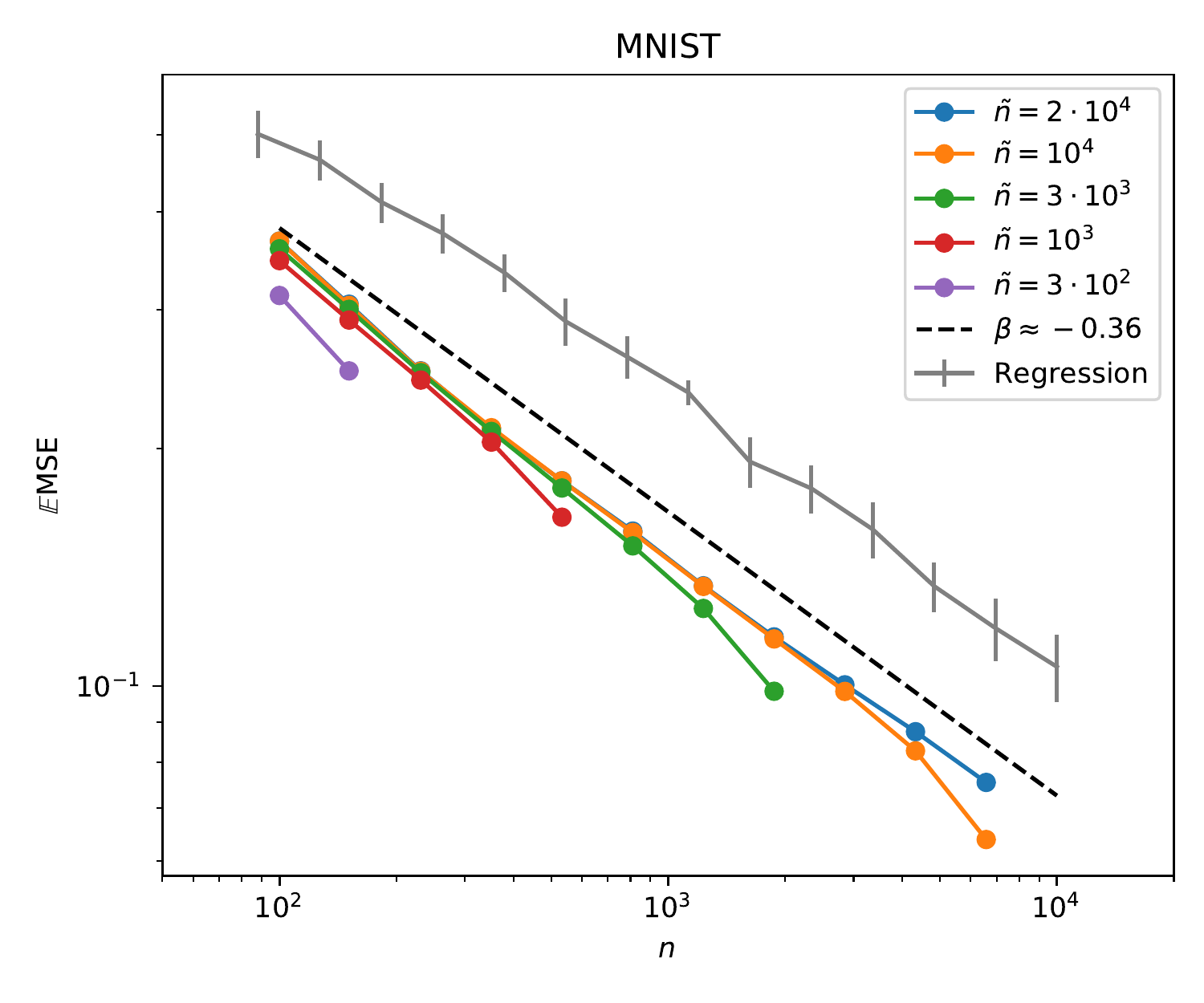}\includegraphics[width=0.5\textwidth]{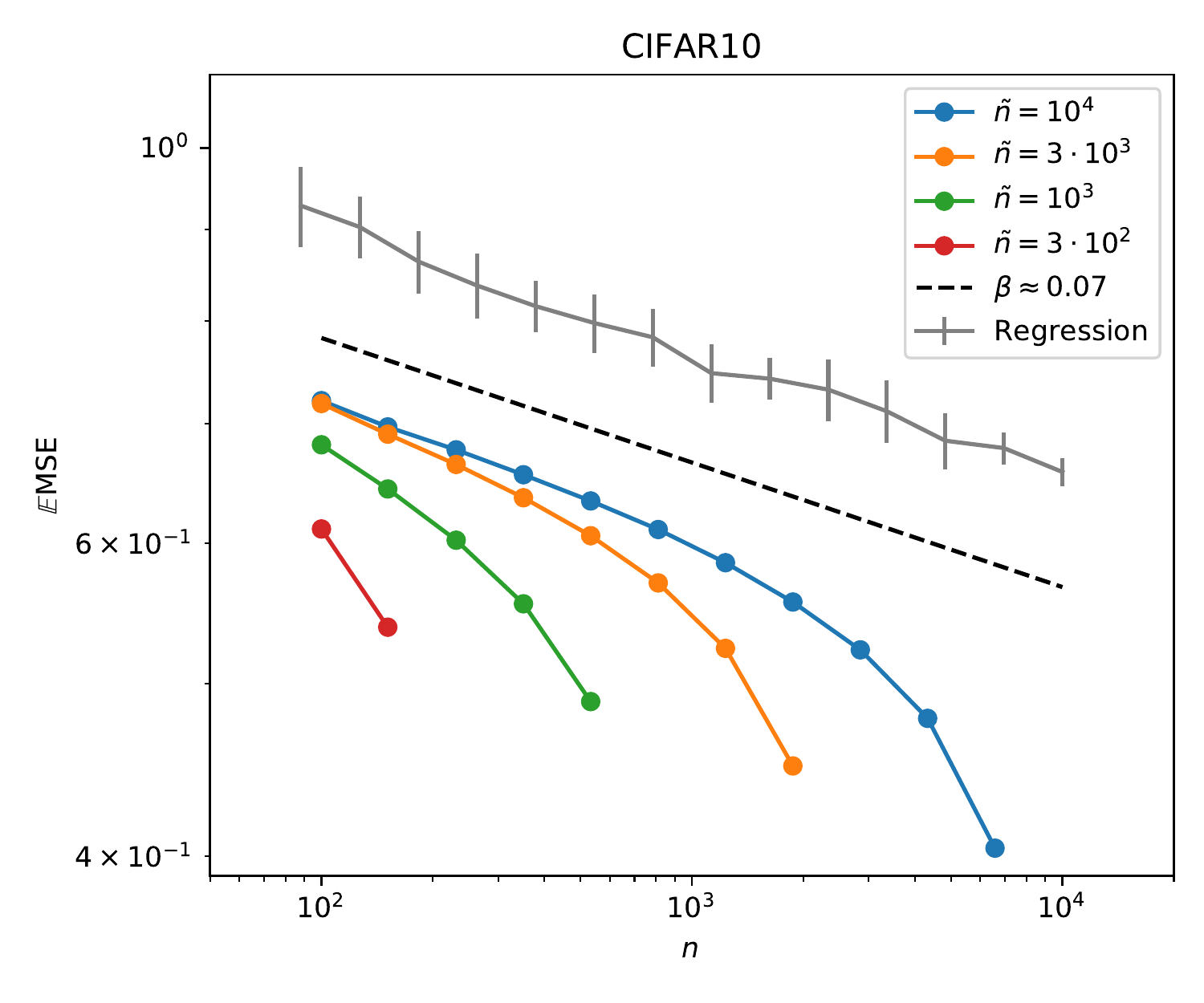}
\caption{Several measures of the learning curves for MNIST (\emph{left}) and CIFAR10 (\emph{right}). In every plot, the gray solid line is the numerical evaluation of the generalization error (shifted for clarity). Colored lines are computed using \eref{eq:realprediction} for several values of $\tl n$, and the dashed black line is the a fit of the power-law decay with which we extract the predicted exponents $\hat\beta_\mr{MNIST} \approx 0.36$ and $\hat\beta_\mr{CIFAR10} \approx 0.07$. \label{fig:realdatapredictederrors}}
\end{figure}

 Finally, we approximate \eref{eq:interpretmodes}  as:
\begin{equation}
    \sum_{\rho=n}^{\tl n} \left(\vl Z \cdot \vl\phi_\rho\right)^2 . \label{eq:realprediction}
\end{equation}
This quantity  is plotted in \fref{fig:realdatapredictederrors}, where we show that it correlates remarkably well with the true learning curve. Fitting these cumulative curves whose exponent is $1-a$ for asymptotically large $\tl n$ (here we plotted several curves for growing $\tl n$) we  extract  an exponent $a_\mr{MNIST}=1.36$ leading to $\hat\beta_\mr{MNIST} \approx 0.36$ and $a_\mr{CIFAR10}=1.07$ leading to $\hat\beta_\mr{CIFAR10} \approx 0.07$ that are very close to the exponents that we measured in \sref{sec:genrealdata}. 

Support for the genericity of \eref{eq:interpretmodes} can be obtained from the recent paper \cite{bordelon2020spectrum}, where the authors derived a formula for the  generalization error based on the decomposition of the target function on the eigenbasis of the kernel. The formula is derived with uncontrolled approximations, but applies to a generic target function (or ensembles thereof) and a generic data point distribution $p(\vl x)$, and matches well their numerical experiments. In \aref{app:approxresult} we show that the asymptotic limit (large $n$) of their formula yields \eref{eq:interpretmodes}. Furthermore, Eq.\ref{eq:maintheorem} is recovered if a power-law decay of the coefficient of the true function in the eigenbasis of the kernel is assumed- thus generalizing our result to non-Gaussian data. 

\section{Effective dimension 
of real data}\label{sec:randomandrealdatascaling}

\begin{figure}[tbh!]
\centering
\includegraphics[width=\textwidth]{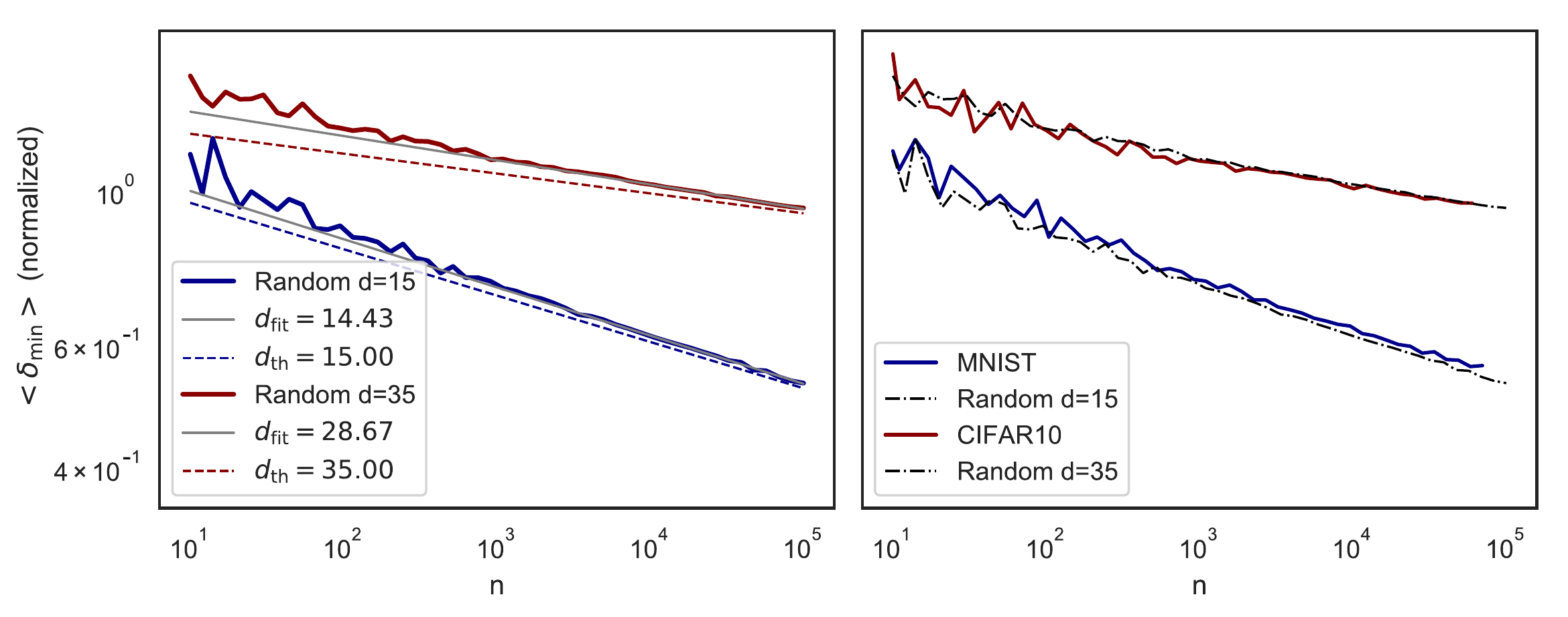}
\caption{Average distance from one point to its nearest neighbor as a function of the dataset size $n$. (\emph{Left}) For random points on $d$-dimensional hypersphere, $\left<\delta_\mathrm{min}\right> \sim n^{-\nicefrac1d}$. Colored solid curves are found numerically, dashed lines are the theoretical asymptotic prediction and the gray lines are numerical fit (we fitted only starting from $n\approx6000$ to reduce finite size effects, and the fit have been rescaled to match the data at $n=10$). The larger $d$, the stronger the preasymptotic effects (a larger $n$ is needed to observe the predicted scaling). (\emph{Right}) Comparison between random data on $15$- and $35$-dimensional hyperspheres and the MNIST, CIFAR10 datasets. According to this definition of effective dimension, MNIST live on a $15$-dimensional manifold and CIFAR10 on a $35$-dimensional one. Data have been rescaled along the $y$-axis for ease of comparison.\label{fig:scaling_delta_min}}
\end{figure}

Both our predictions and empirical observations support rather large values of $\beta$. From a Gaussian random process point of view, it is surprising:  the exponent $\beta$ avoids the curse of dimensionality only if the smoothness of the Teacher is of the order of the data dimension. If these observations were to hold true also for real data, it would seem to imply that MNIST and CIFAR10 must be hundreds or thousands of times differentiable! However, there is a simple catch: real data actually live on a manifold of much lower dimensionality. Above we have argued that the quantity that governs the asymptotic learning curve is the typical distance $\delta_\mr{min}$ between neighboring points in the training set. A simple way to measure the effective dimension $d_\mr{eff}$ of real data consists then in plotting the (asymptotic) dependence of $\delta_\mr{min}$ on the number of points $n$ in a random subset of the dataset, and fitting
\begin{equation}
    \delta_\mr{min}\sim n^{-\nicefrac1{d_\mr{eff}}}.
\end{equation}
In \fref{fig:scaling_delta_min} (\emph{right}) we show that for MNIST and CIFAR10 there is indeed a power-law relation linking $\delta_\mr{min}$ to $n$, and that the effective dimension extracted this way is much smaller than the embedding dimension of the datasets:
\begin{align}
    d_\mr{eff}^\mr{MNIST} \approx 15 &\ll 784 = d^\mr{MNIST},\\
    d_\mr{eff}^\mr{CIFAR10} \approx 35 &\ll 3072 = d^\mr{CIFAR10}.
\end{align}
This  measure is consistent with previous extrapolations of the intrinsic dimension of MNIST~\cite{costa2004learning,hein2005intrinsic,rozza2012novel,facco2017estimating}.


\section{Conclusion}

In this work we have shown for CIFAR10 and MNIST  respectively that kernel regression and classification display a power-law decay in the learning curves, quite remarkably with essentially the same exponent $\beta$ regardless of task and kernel --- a fact yet to be explained. These exponents are much larger than $\beta=1/d$ expected for Lipschitz target functions and smaller than $\beta=1/2$ expected for RKHS target functions. 

This observation led us to study a Teacher-Student framework for regression in which data are modeled as Gaussian random fields of varying smoothness, in which intermediary values of $\beta$ are obtained. We find two regimes depending on the respective smoothness of the Teacher and Student kernels. If the student is smooth enough --- i.e. it puts a sufficiently low prior on high  frequency components --- then $\beta$ is entirely controlled by the Teacher. We obtain that the smoothness index must scale with the dimension for $\beta$ to be finite as $d\rightarrow \infty$, recovering the curse of dimensionality. 

In our calculations, the dimension enters as the parameter relating the number of points to the nearest-neighbor  distance $\delta\sim n^{-1/d}$. Thus in practice the parameter $d$ considered should be the effective dimension $d_{eff}$ of the data, which is much smaller than the number of pixels for MNIST and CIFAR data. It explains why $\beta$ is not very small in these cases. 

Finally, for Gaussian fields our result is equivalent to the statement that $\beta$ is governed by the power of the true function past the first $\sim n$ eigenvectors of the kernel. We test this more general idea both for CIFAR and MNIST and find that it correctly predicts $\beta$. Understanding what controls this power in a general setting (which include the effective dimension of the data and presumably a generalized quantity characterizing smoothness) thus appears necessary to understand how many data are required to learn a task. 



\subsubsection*{Acknowledgments}
We acknowledge G.~Biroli, C.~Hongler and F.~Gabriel for the discussions that stimulated this work, and we thank S.~d'Ascoli, A.~Jacot,   C. Pehlevan,  L.~Sagun and M.L.~Stein for discussions. 
This work was partially supported by the grant from the Simons Foundation (\#454953 Matthieu Wyart). M.W. thanks the Swiss National Science Foundation for support under Grant No.~200021-165509.

\bibliographystyle{unsrt}
\bibliography{main}{}
\clearpage
\appendix

\section{Soft-margin Support Vector Machines}\label{app:smsvms}

The kernel classification task is performed via the algorithm known as \emph{soft-margin Support Vector Machine}.

We want to find a function $\hat Z_S(\vl x)$ such that its sign correctly predicts the label of the data. In this context we model such a function as a linear prediction after projecting the data on a \emph{feature space} via $\vl x \to \phi(\vl x)$:
\begin{equation}
\hat Z_S(\vl x) = \vl w \cdot \vl\phi_S(\vl x_\mu) + b,
\end{equation}
where $\vl w,b$ are parameters to be learned from the training data. The kernel is related to the feature space via $K_S(\vl x, \vl x^\prime) = \vl\phi_S(\vl x)\cdot\vl\phi_S(\vl x^\prime)$.  We require that $Z(\vl x_\mu) \hat Z_S(\vl x_\mu) > 1 - \xi_\mu$ for all training points. Ideally we want to have some large margins $1-\xi_\mu=1$, but we allow some of them to be smaller by introducing the \emph{slack variables} $\xi_\mu$ and penalizing large values. To achieve this the following constrained minimization is performed:
\begin{equation}
\min_{\vl w,b,\vl \xi} \frac12 \abs{\vl w}^2 + C \sum_\mu \xi_\mu \quad \mr{subjected\ to} \quad \forall\mu \ Z(\vl x_\mu) \left[\vl w\cdot\vl\phi_S(\vl x_\mu) + b\right] \geq 1-\xi_\mu, \ \xi_\mu\geq0.
\end{equation}
This problem can be expressed in a dual formulation as
\begin{equation}
\min_{\vl a} \frac12 \vl a\cdot \mb Q_S\vl a - \sum_{\mu=1}^n a_\mu \quad \mr{subjected\ to} \quad \vl Z\cdot\vl a = 0,\ 0\leq a_\mu\leq C,
\end{equation}
where $\mb Q_{S,\mu,\nu} = Z(\vl x_\mu)Z(\vl x_\nu) K_S(\vl x_\mu, \vl x_\nu)$ and $\vl Z$ is the vector of the labels of the training points. Here $C$ ($=10^4$ in our simulations) controls the trade-off between minimizing the training error and maximizing the \emph{margins} $1-\xi_\mu$. For the details we refer to~\cite{scholkopf2001learning}. If $\vl a^\star$ is the solution to the minimization problem, than
\begin{align}
&\vl w^\star = \sum_\mu a_\mu^\star \vl\phi_S(\vl x_\mu),\\
&b^\star = Z(\vl x_\mu) - \sum_\nu a_\nu y_\nu K_S(\vl x_\mu, \vl x_\nu) \quad \mr{for\ any}\ \mu\ \mr{such\ that}\ a_\mu<C.
\end{align}
The predicted label for unseen data points is then
\begin{equation}
\mr{sign}(\hat Z_S(\vl x)) = \mr{sign}(\sum_\mu Z(\vl x_\mu) a_\mu K_S(\vl x_\mu, \vl x) + b^\star)
\end{equation}
The generalization error is now defined as the probability that an unseen image has a predicted label different from the true one, and such a probability is again estimated as an average over a test set with $n_\mr{test}$ elements:
\begin{equation}
\mr{Error} = \frac1{n_\mr{test}}\sum_{\mu=1}^{n_\mr{test}} \theta\left[-\mr{sign}\left(\hat Z_S(\vl x_\mu)\right) Z(\vl x_\mu)\right].
\end{equation}

\vspace{2em}
\section{Different kernel variances}\label{app:diffsigmas}

\begin{figure}[t]
\centering
\includegraphics[width=0.6\textwidth]{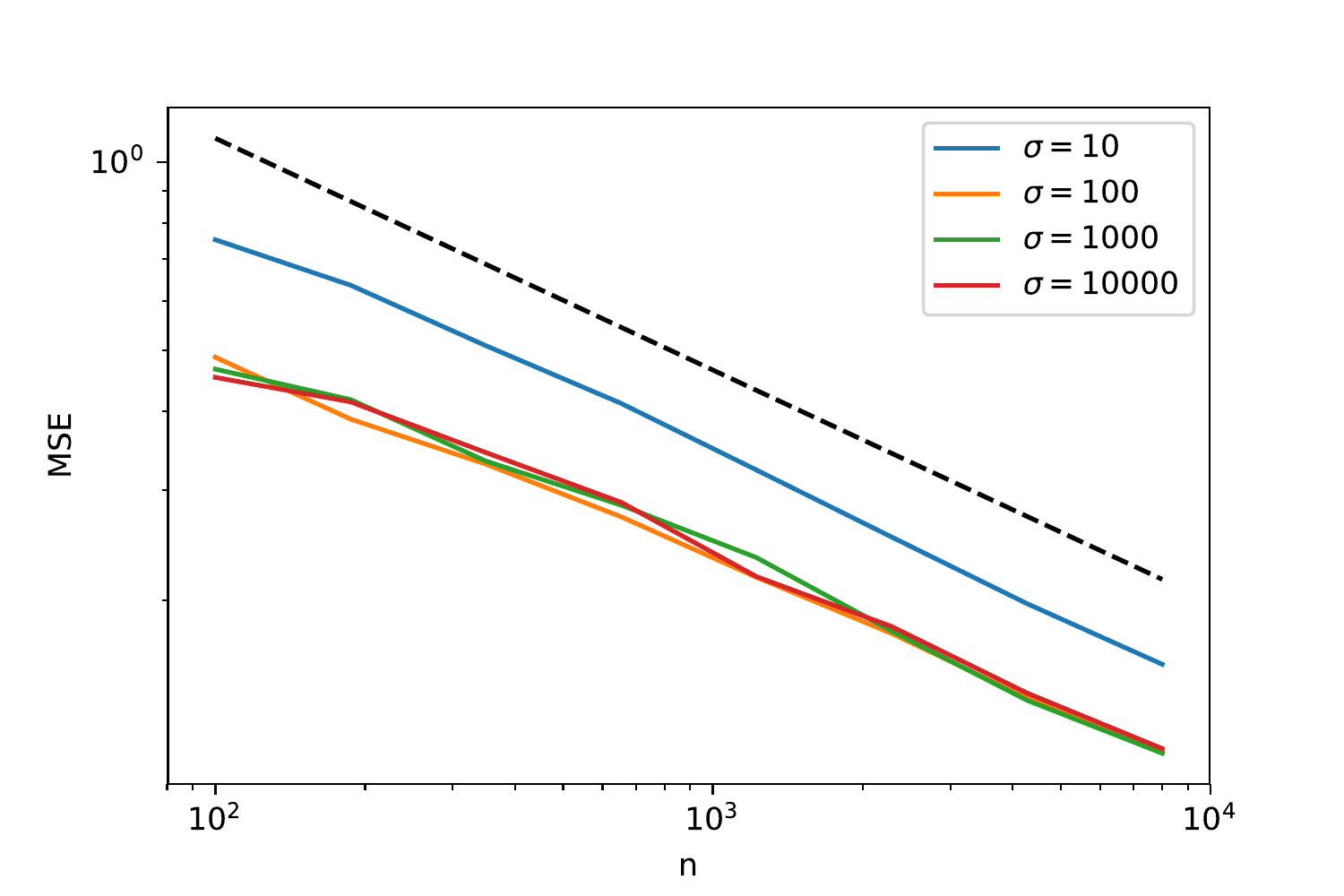}
\caption{Learning curves for kernel regression on the MNIST dataset. Regression is performed with several Laplace kernels of varying variance $\sigma$ ranging from $\sigma=10$ to $\sigma=10000$. \label{fig:diff_sigmas}}
\end{figure}

In \fref{fig:diff_sigmas} we show the learning curves for kernel regression on the MNIST (parity) dataset --- the same setting as in \fref{fig:realdata} \emph{(a)}. Several Laplace kernels of varying variance $\sigma$ are used. The variance ranges several orders of magnitude and the learning curves all decay with the same exponent, although for $\sigma=10$ the algorithm achieves suboptimal performance and the test errors are increased by some factor.

\section{Different choice of kernel variances}\label{app:othersigmas}

\begin{figure}[b!]
\centering
\includegraphics[width=0.5\textwidth]{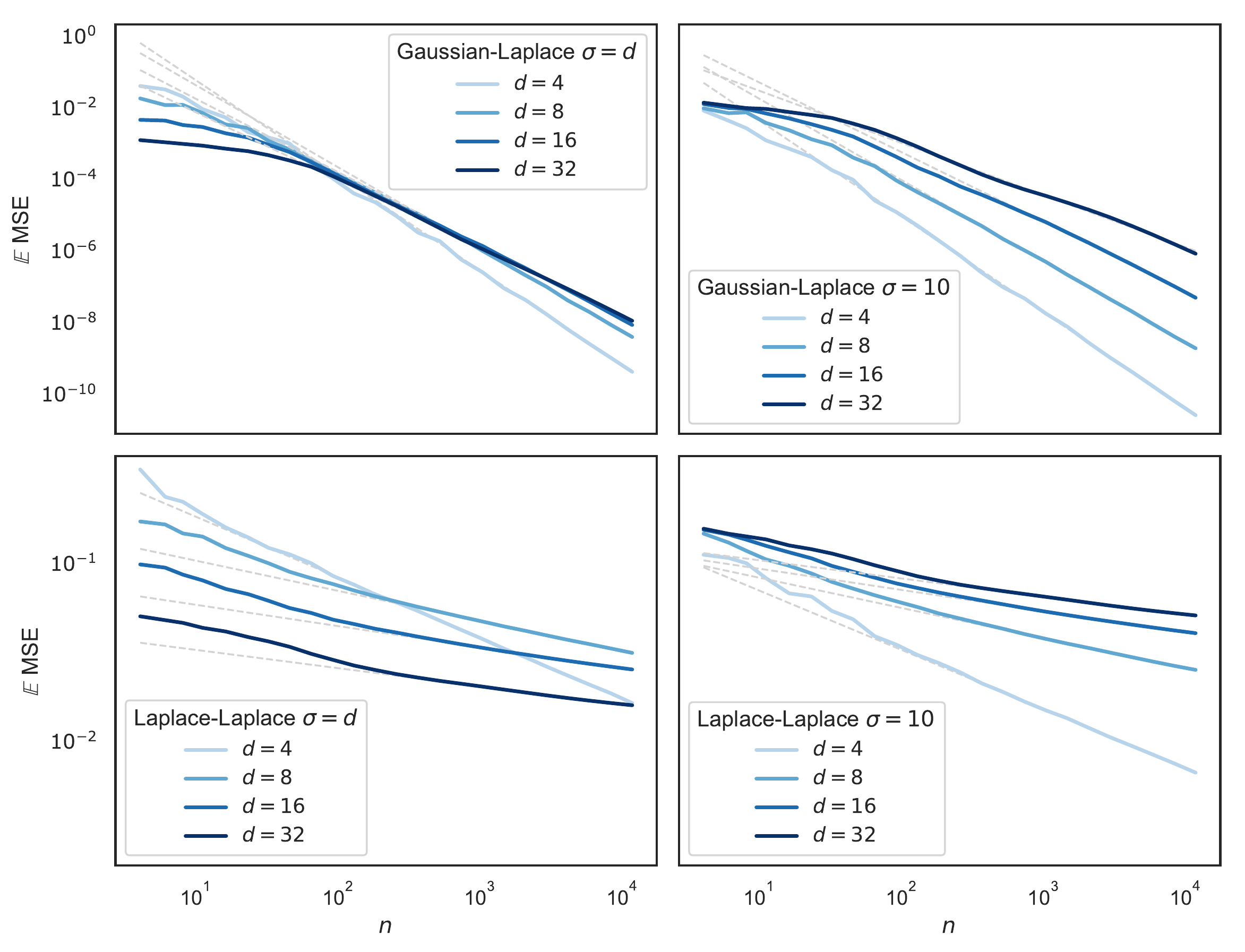}\includegraphics[width=0.5\textwidth]{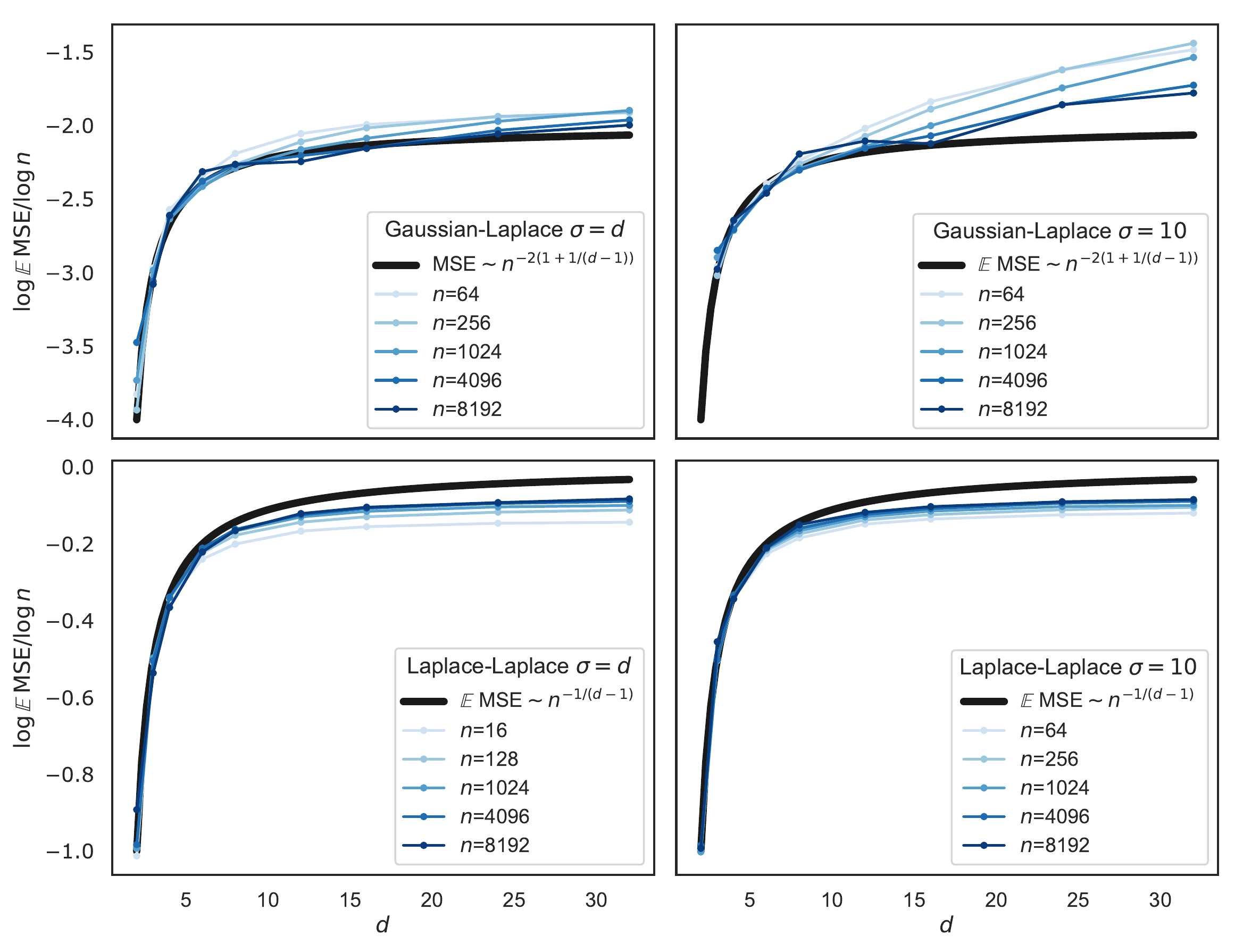}
\caption{In these plots we show the results for the Teacher-Student kernel regression. The Student is always a Laplace kernel, the Teacher is either Gaussian or Laplace. The four plots on the left depict the mean-square error against the size of the dataset for different spatial dimensions of the data, those on the right show the fitted asymptotic exponent against the spatial dimension for different dataset sizes. For every case we show both the the results for $\sigma_T=\sigma_S=d$ and $\sigma_T=\sigma_S=10$.\label{fig:appothersigmas}}
\end{figure}

In \fref{fig:appothersigmas} we show the learning curves for the Teacher-Student kernel regression problem, with a Student kernel that is always Laplace and a Teacher that can be either Gaussian or Laplace. We show how the test error decays with the size of the training dataset and how the asymptotic exponent depends on the spatial dimension. Every experiment is run with two different choices of the kernel variances: in one case $\sigma_T=\sigma_S=d$ and in the other $\sigma_T=\sigma_S=10$. We observed that scaling the variances with the spatial dimension leads faster to the results that we predicted in this paper, but overall the choice has little effect on the exponents (both tend towards the prediction as the dataset size is increased).

\begin{figure}[t!]
\centering
\includegraphics[width=0.33\textwidth]{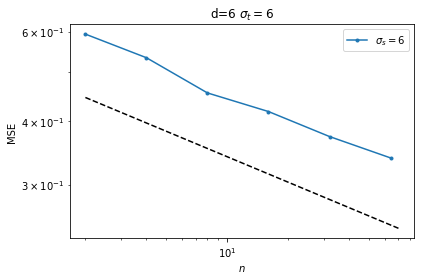}\includegraphics[width=0.34\textwidth]{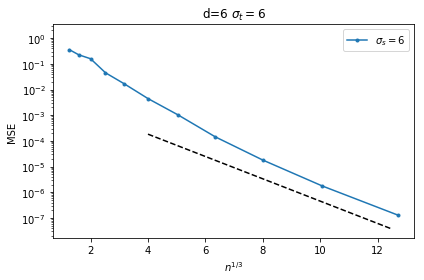}\includegraphics[width=0.33\textwidth]{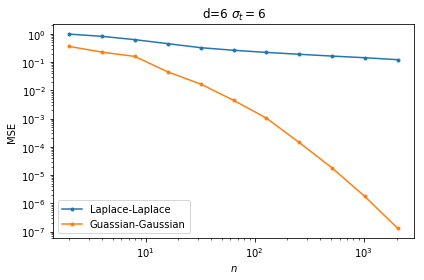}
\caption{{\bf Left:} The test error of a Laplace Teacher ($\alpha_T=d+1$) with a Gaussian Student ($\alpha_S=\infty$) decays as a power law with the predicted exponent $\beta = \frac1d \min(1, \infty) = \frac16$ in $d=6$ dimensions. {\bf Center:} When both the Teacher and the Student are Gaussian the test error decays faster than any power law as the number $n$ of data is increased. This plot confirm this by showing that the logarithm of the test error decays linearly as a function of $n^\frac13$. {\bf Right:} Comparison between the learning curves for the cases where both kernels are either Laplace (top blue line) or Gaussian (bottom orange line). While the former decays algebraically with the predicted exponent, the latter decays exponentially, in agreement with the prediction $\beta = \infty$ found within our framework.   In all these plots we have taken the variances of both the Teacher and Student kernels to be equal to the dimension $d=6$. \label{fig:gaussianstudent}}
\end{figure}

\section{Gaussian Students}\label{app:gaussianstudent}

In this appendix we present the learning curves of Gaussian Students: the Fourier transform of these kernels decays faster than any power law and one can effectively consider $\alpha_S=\infty$. If the Teacher is Laplace ($\alpha_T=d+1$) then the predicted exponent is finite and takes the values $\beta = \frac1d \min(\alpha_T-d, 2\alpha_S) = \frac1d \min(1, \infty) = \frac1d$. Such a case is displayed in \fref{fig:gaussianstudent} \emph{(left)} in dimension $d=6$. However, if we consider the Teacher to be Gaussian as well, then the predicted exponent would be $\beta = \frac1d\min(\infty, \infty) = \infty$. This case corresponds to \fref{fig:gaussianstudent} \emph{(center)}: the test errors decays faster than a power law. In \fref{fig:gaussianstudent} \emph{(right)} we compare the case where both kernels are Gaussian to the case where both kernels are Laplace: while the latter decays as a power law, the former decays much faster.

\section{Mat\'ern Teachers}\label{app:matern}

To further test the applicability of our theory, we show here some numerical simulations for a Teacher kernel that is a Mat\'ern covariance function and a Laplace kernel as student. We ran the simulations in 1d: the data points are sampled uniformly on a 1-dimensional circle embedded in $\mb R^2$. Mat\'ern kernels are parametrized by a parameter $\nu>0$:
\begin{equation}
    K_T(\vl x) = \frac{2^{1-\nu}}{\Gamma(\nu)} z^\nu \mc K_\nu(z),
\end{equation}
where $z=\sqrt{2\nu} \frac{\abs{\vl x}}{\sigma}$ ($\sigma$ being the kernel variance), $\Gamma$ is the gamma function and $\mc K_\nu$ is the Bessel function of the second kind with parameter $\nu$. Interestingly we recover the Laplace kernel for $\nu=\nicefrac12$ and the Gaussian kernel for $\nu=\infty$. As one can find in e.g. \cite{williams2006gaussian}, the exponent $\alpha_T$ that governs the decay at high frequency of this kernels is $\alpha_T=d+2\nu$. Varying $\nu$ we can change the smoothness of the target function.

For $d=1$ our prediction for the learning curve exponent $\beta$ is
\begin{equation}
    \beta = \frac1d \min(\alpha_T-d,2\alpha_S) = \min(2\nu,4).
\end{equation}

\begin{figure}[t!]
\centering
\includegraphics[width=0.8\textwidth]{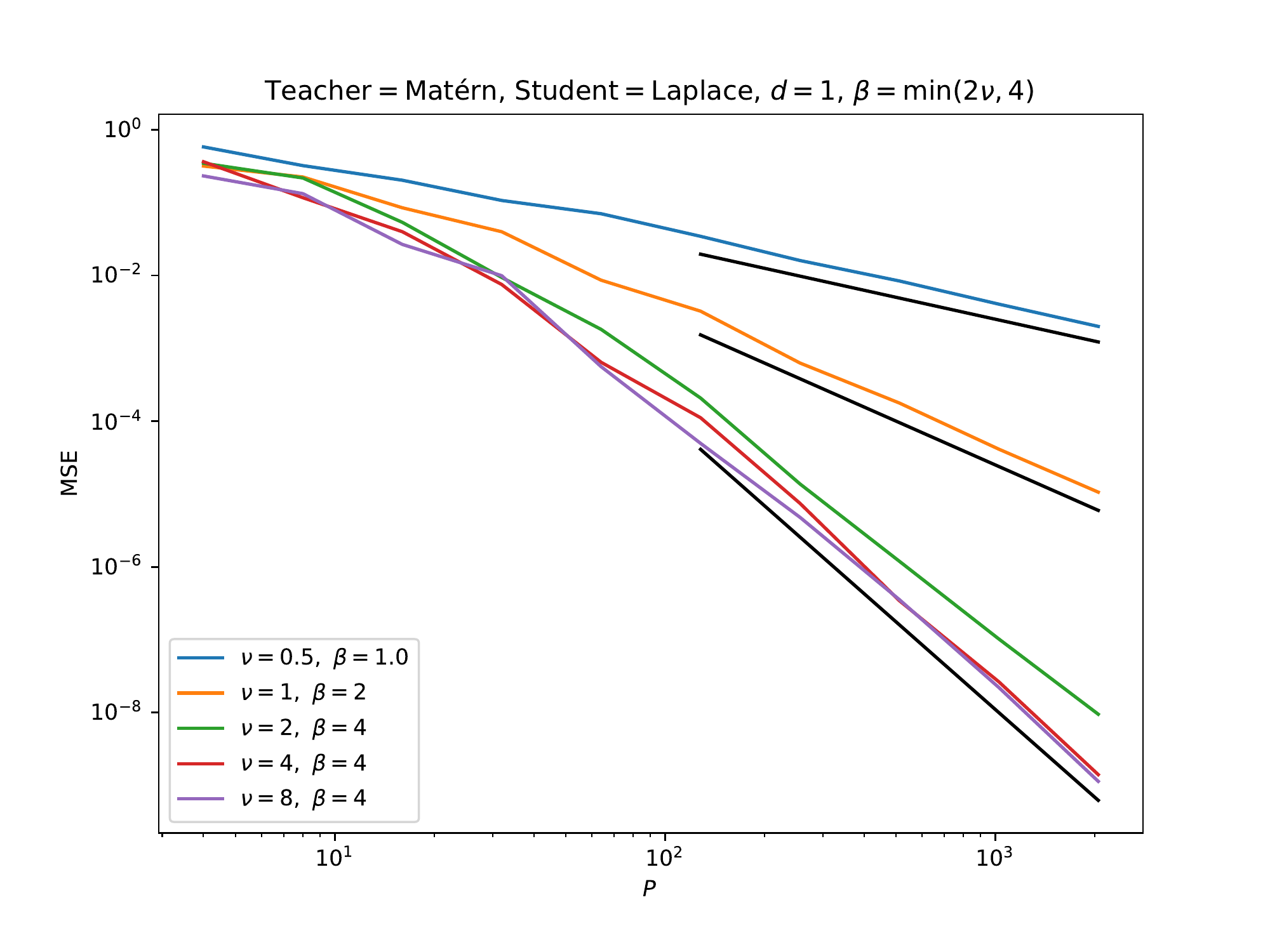}
\caption{Mean-squared error for Mat\'ern Teacher kernels and Laplace students. The variance of the kernels is equal to $2$ for all the curves.\label{fig:matern}}
\end{figure}

In \fref{fig:matern} we verify that our prediction matches the numerical results.

\section{Proof of theorem}\label{app:maintheorem}

We prove here Theorem \ref{th:theorem}:

\textbf{Theorem 1} \emph{Let $\tl K_T(\vl w) = c_T \abs{\vl w}^{-\alpha_T} + o\left(\abs{\vl w}^{-\alpha_T}\right)$ and $\tl K_S(\vl w) = c_S \abs{\vl w}^{-\alpha_S} + o\left(\abs{\vl w}^{-\alpha_S}\right)$ as $\abs{\vl w}\to\infty$, where $\tl K_T(\vl w)$ and $\tl K_S(\vl w)$ are the Fourier transforms of the kernels $K_T(\vl x)$, $K_S(\vl x)$ respectively, assumed to be positive definite. We assume that $\tl K_T(\vl w)$ and $\tl K_S(\vl w)$ have a finite limit as $\abs{\vl w}\to0$ and that $K(\vl 0)<\infty$. Then,
\begin{equation}
\mb E\,\mr{MSE} = n^{-\beta} + o\left(n^{-\beta}\right) \quad \mr{with} \quad \beta=\frac1d\min(\alpha_T-d, 2\alpha_S).
\end{equation}
Moreover, in the case of a Gaussian kernel the result holds valid if we take the corresponding exponent to be $\alpha=\infty$.}

\begin{proof}
Our strategy is to compute how the mean-square test error scales with distance $\delta$ between two nearest neighbors on the $d$-dimensional regular lattice. At the end, we will use the fact that $\delta \propto n^{-\nicefrac1d}$, where $n$ is the number of sampled points on the lattice.

We denote by $\tl F(\vl w)$ the Fourier transform of a function $F:\mc V\to\mb R$:
\begin{align}
&\tl F(\vl w) = L^{-\nicefrac{d}2} \int_{\mc V} \mr{d}\vl x\, e^{-i\vl w\cdot\vl x} F(\vl x), \quad \mr{where}\ \vl w \in \mb{L}\equiv\frac{2\pi}L \mb{Z}^d,\\
&F(\vl x) = L^{-\nicefrac{d}2} \sum_{\vl w\in\mb{L}} e^{i\vl w\cdot\vl x} \tl F(\vl w).
\end{align}
If $Z \sim \mc{N}(0, K)$ is a Gaussian field with translation-invariant covariance $K$ then by definition
\begin{equation}
\mb E Z(\vl x) Z(\vl x^\prime) = K(\vl x - \vl x^\prime).
\end{equation}

\emph{Properties of the Fourier transform of a Gaussian field:}
\begin{align}
&\tl K(\vl w) = \tl K(-\vl w) \in \mb R,\label{eq:ftgf1}\\
&\mb E \tl Z(\vl w) = 0,\label{eq:ftgf2}\\
&\mb E \tl Z(\vl w) \overline{\tl Z(\vl w^\prime)} = L^{\nicefrac{d}2} \delta_{\vl w \vl w^\prime} \tl K(\vl w).\label{eq:ftgf3}
\end{align}
\eref{eq:ftgf1} comes from the fact that $K(\vl x)$ is an even, real-valued function. The real and imaginary parts of $\tl Z(\vl w)$ are Gaussian random variables. They are all independent except that $\tl Z(-\vl w) = \overline{\tl Z(\vl w)}$. \eref{eq:ftgf3} follows from the fact that $Z(\vl x)$ and $K(\vl x)$ are $L$-periodic functions, and therefore $e^{i\vl w\cdot\vl x} \tl K(\vl w)$ is the Fourier transform of $K(\cdot + \vl x)$ if $\vl w\in\frac{2\pi}L\mb Z^d$. \hfill{\#\quad}

The solution \eref{eq:regrpred} for kernel regression has two interpretations. In \sref{sec:gentsproblem} we introduced it as the quantity that minimizes a quadratic error, but it can also be seen as the \emph{maximum-a-posteriori} (MAP) estimation of another formulation of the problem~\cite{williams2006gaussian}. The field $Z(\vl x)$ is assumed to be drawn from a Gaussian distribution with covariance function $K_S(\vl x)$: $K_S$ therefore plays a role in the \emph{prior} distribution of the data $\vl Z = (Z(\vl x_\mu)_{\mu=1}^n)$. Inference about the value of the field $\hat Z_S(\vl x)$ at another location is then performed by maximizing its posterior distribution,
\begin{equation}
\hat Z_S(\vl x) \equiv \mathrm{arg\ max}\ \mc P\left( Z(\vl x) | \vl Z \right).
\end{equation}
Such a posterior distribution is Gaussian, and its mean --- and therefore also the value that maximizes the probability --- is exactly \eref{eq:regrpred}:
\begin{equation}
\hat Z_S(\vl x) = \vl k_S(\vl x) \cdot \mb K_S^{-1} \vl Z,
\end{equation}
where where $\vl Z = \left(Z(\vl x_\mu)\right)_{\mu=1}^n$ are the training data, $\vl k_S(\vl x) = \left(K_S(\vl x_\mu, \vl x)\right)_{\mu=1}^n$ and $\mb K_S = \left(K_S(\vl x_\mu, \vl x_\nu)\right)_{\mu,\nu=1}^n$ is the Gram matrix, that is invertible since the kernel $K_S$ is assumed to be positive definite. By Fourier transforming this relation we find
\begin{equation}
\tl Z_S(\vl w) = \tl Z^\star(\vl w) \frac{\tl K_S(\vl w)}{\tl K_S^\star(\vl w)},\label{eq:zpost}
\end{equation}
where we have defined $F^\star(\vl w) \equiv \sum_{\vl n\in\mb Z^d} F\left(\vl w + \frac{2\pi\vl n}\delta\right)$ for a generic function $F$.

Another way to reach \eref{eq:zpost} is to consider that we are observing the quantities
\begin{equation}
\tl Z^\star(\vl w) \equiv \delta^d L^{-\nicefrac{d}2} \sum_{\vl x\in\mr{lattice}} e^{-i\vl w\cdot\vl x}Z(\vl x) \equiv \sum_{\vl n\in\mb Z^d} \tl Z\left(\vl w + \frac{2\pi\vl n}\delta\right).\label{eq:bayessum}
\end{equation}
Given that we know the prior distribution of the Fourier components on the right-hand side in \eref{eq:bayessum}, we can infer their posterior distribution once their sums are constrained by the value of $\tl Z^\star(\vl w)$, and it is straightforward to see that we recover \eref{eq:zpost}.

The mean-square error can then be written using the Parseval-Plancherel identity,
\begin{equation}
\mb E\,\mr{MSE} = L^{-d} \mb E\int_{\mc V} \mr{d}\vl x\, [Z(\vl x) - \hat Z_S(\vl x)]^2 = L^{-d} \mb E \sum_{\vl w\in \mb L} \left|\tl Z(\vl w) - \tl Z^\star(\vl w) \frac{\tl K_S(\vl w)}{\tl K_S^\star(\vl w)}\right|^2.
\end{equation}

By taking the expectation value with respect to the Teacher and using \eref{eq:ftgf1}-\eref{eq:ftgf3} we can write the mean-square error as
\begin{multline}
\mathbb E\,\mr{MSE} = L^{-d} \mb E \sum_{\vl w\in\mb L} \left[ \tl Z(\vl w) \overline{\tl Z(\vl w)} - 2 \tl Z(\vl w) \overline{\tl Z^\star(\vl w)} \frac{\tl K_S(\vl w)}{\tl K_S^\star(\vl w)} + \tl Z^\star(\vl w) \overline{\tl Z^\star(\vl w)} \frac{\tl K_S^2(\vl w)}{{\tl K_S^\star{}}^{2}(\vl w)} \right] = \\
= L^{-d} \mb E \sum_{\vl w\in\mb L} \tl Z(\vl w)\overline{\tl Z(\vl w)} - 2\frac{\tl K_S(\vl w)}{\tl K_S^\star(\vl w)} \sum_{\vl n\in\mb Z^d} \tl Z(\vl w) \overline{\tl Z\left(\vl w + \frac{2\pi\vl n}{\delta}\right)} + \\
+ \frac{\tl K_S^2(\vl w)}{{\tl K_S^\star{}}^{2}(\vl w)} \sum_{\vl n,\vl n^\prime\in\mb Z^d} \tl Z\left(\vl w + \frac{2\pi\vl n}{\delta}\right) \overline{\tl Z\left(\vl w + \frac{2\pi\vl n^\prime}{\delta}\right)} = \\
= L^{-\nicefrac{d}{2}} \sum_{\vl w\in\mb L} \tl K_T(\vl w) - 2\frac{\tl K_S(\vl w)}{\tl K_S^\star(\vl w)} \tl K_T(\vl w) + \frac{\tl K_S^2(\vl w)}{{\tl K_S^\star{}}^{2}(\vl w)} \sum_{\vl n\in\mb Z^d} \tl K_T\left(\vl w + \frac{2\pi\vl n}{\delta}\right) = \\
= L^{-\nicefrac{d}2} \sum_{\vl w \in\mb L\cap\mc B} \tl K_T^\star(\vl w) - 2 \frac{[\tl K_T\tl K_S]^\star(\vl w)}{\tl K_S^\star(\vl w)} + \frac{\tl K_T^\star(\vl w) [\tl K_S^2]^\star(\vl w)}{\tl K_S^\star(\vl w)^2},\label{eq:mseprefinal}
\end{multline}
where $\mathcal B = \left[-\frac\pi\delta, \frac\pi\delta\right]^d$ is the Brillouin zone.

At high frequencies, $\tl K_T(\vl w) = c_T\abs{\vl w}^{-\alpha_T} + o\left(\abs{\vl w}^{-\alpha_T}\right)$ and $\tl K_S(\vl w) = c_S\abs{\vl w}^{-\alpha_S} + o\left(\abs{\vl w}^{-\alpha_S}\right)$. Therefore:
\begin{multline}
\tl K_T^\star(\vl w) = \tl K_T(\vl w) + \delta^{\alpha_T} c_T \!\!\! \sum_{\vl n\in\mb Z^d\setminus\{\vl 0\}} \!\! \abs{\vl w \delta + 2\pi\vl n}^{-\alpha_T} + o\left(\abs{\vl w}^{-\alpha_T}\right) \equiv \\
\equiv \tl K_T(\vl w) + \delta^{\alpha_T} c_T \, \psi_T(\vl w \delta) + o\left(\abs{\vl w}^{-\alpha_T}\right).\label{eq:psiexp}
\end{multline}
This equation defines the function $\psi_T$, and a similar equation holds for the Student as well.
The hypothesis $K_T(\vl 0) \propto \int\mr{d}\vl w\, \tl K_T(\vl w)<\infty$ implies $\alpha_T > d$ and therefore $\sum_{\vl n\in \mb Z^d} \abs{\vl n}^{-\alpha_T} < \infty$ (and likewise for the Student). Then, $\psi_{\alpha_T}(\vl 0), \psi_{\alpha_S}(\vl 0)$ are finite; furthermore, the $\vl w$'s in the sum \eref{eq:mseprefinal} are at most of order $\mc O\left(\delta^{-1}\right)$, therefore the terms $\psi_\alpha(\vl w\delta)$ are $\mc O(\delta^0)$ and do not influence how \eref{eq:mseprefinal} scales with $\delta$. Applying \eref{eq:psiexp}, expanding for $\delta\ll1$ and keeping only the leading orders, we find
\begin{multline}
\mathbb E\,\mr{MSE} = \\
= L^{-\nicefrac{d}2} \left[ \sum_{\vl w \in\mb L\cap\mc B} 2c_T\psi_{\alpha_T}(\vl w\delta) \delta^{\alpha_T} + c_S^2\psi_{2\alpha_S}(\vl w\delta) \frac{\tl K_T(\vl w)}{\tl K_S^2(\vl w)} \delta^{2\alpha_S} + o\left(\abs{\vl w}^{-\alpha_T}\right) + o\left(\abs{\vl w}^{-2\alpha_S}\right) \right] = \\
= L^{-\nicefrac{d}2} \left[ \sum_{\vl w \in\mb L\cap\mc B} 2c_T\psi_{\alpha_T}(\vl w\delta) \delta^{\alpha_T} + c_S^2\psi_{2\alpha_S}(\vl w\delta) \frac{\tl K_T(\vl w)}{\tl K_S^2(\vl w)} \delta^{2\alpha_S}\right] + o\left(\abs{\vl w}^{-\alpha_T - d}\right) + o\left(\abs{\vl w}^{-2\alpha_S - d}\right).\label{eq:scalingsum}
\end{multline}
We have neglected terms proportional to, for instance, $\delta^{\alpha_T+\alpha_S}$, since they are subleading with respect to $\delta^{\alpha_T}$, but we must keep both $\delta^{\alpha_T}$ and $\delta^{\alpha_S}$ since we do not know a priori which one is dominant. The additional term $\delta^{-d}$ in the subleading terms comes from the fact that $|\mb L \cap \mc B| = \mc O\left(\delta^{-d}\right)$.

The first term in \eref{eq:scalingsum} is the simplest to deal with: since $\abs{\vl w\delta}$ is smaller than some constant for all $\vl w \in \mb L\cap\mc B$ and the function $\psi_{\alpha_T}(\vl w\delta)$ has a finite limit, we have
\begin{equation}
\delta^{\alpha_T} \sum_{\vl w\in\mb L\cap\mc B} 2c_T\psi_{\alpha_T}(\vl w\delta) = \mc O\left(\delta^{\alpha_T} |\mb L\cap \mc B|\right) = \mc O \left(\delta^{\alpha_T-d}\right).\label{eq:scalingfirstterm}
\end{equation}

We then split the second term in \eref{eq:scalingsum} in two contributions:

\paragraph{Small $\abs{\vl w}$} We consider ``small'' all the terms $\vl w\in\mb L\cap\mc B$ such that $\abs{\vl w} < \Gamma$, where $\Gamma \gg1$ is $\mc O(\delta^0)$ but large. As $\delta\to0$, $\psi_{2\alpha_S}(\vl w\delta) \to \psi_{2\alpha_S}(\vl 0)$ which is finite because $K(\vl 0)<\infty$. Therefore
\begin{equation}
\delta^{2\alpha_S} \sum_{\substack{\vl w \in\mb L\cap\mc B\\\abs{\vl w}<\Gamma}} c_S^2\psi_{2\alpha_S}(\vl w\delta) \frac{\tl K_T(\vl w)}{\tl K_S^2(\vl w)} \to \delta^{2\alpha_S} c_S^2\psi_{2\alpha_S}(\vl 0) \sum_{\substack{\vl w \in\mb L\cap\mc B\\\abs{\vl w}<\Gamma}} \frac{\tl K_T(\vl w)}{\tl K_S^2(\vl w)}.
\end{equation}
The summand is real and strictly positive because the positive definiteness of the kernels implies that their Fourier transforms are strictly positive. Moreover, as $\delta\to0$, $\mb L \cap \mc B \cap \{\abs{\vl w} < \Gamma\} \to \mb L \cap \{\abs{\vl w} < \Gamma\}$, which contains a finite number of elements, independent of $\delta$. Therefore
\begin{equation}
\delta^{2\alpha_S} \sum_{\substack{\vl w \in\mb L\cap\mc B\\\abs{\vl w}<\Gamma}} c_S^2\psi_{2\alpha_S}(\vl w\delta) \frac{\tl K_T(\vl w)}{\tl K_S^2(\vl w)} = \mc O\left(\delta^{2\alpha_S}\right).\label{eq:scalingsecondterm1}
\end{equation}


\paragraph{Large $\abs{\vl w}$} ``Large'' $\vl w$ are those with $\abs{\vl w} > \Gamma$: we recall that $\Gamma\gg1$ is $\mc O(\delta^0)$ but large. This allows us to approximate $\tl K_T$, $\tl K_S$ in the sum with their asymptotic behavior:
\begin{multline}
\delta^{2\alpha_S} \sum_{\substack{\vl w \in\mb L\cap\mc B\\\abs{\vl w}>\Gamma}} c_S^2\psi_{2\alpha_S}(\vl w\delta) \frac{\tl K_T(\vl w)}{\tl K_S^2(\vl w)} \propto \delta^{2\alpha_S} \sum_{\substack{\vl w \in\mb L\cap\mc B\\\abs{\vl w}>\Gamma}} \abs{\vl w}^{-\alpha_T+2\alpha_S} + o\left(\abs{\vl w}^{-\alpha_T + 2\alpha_S}\right) \approx \\
\approx \delta^{2\alpha_S} \int_{\Gamma}^{\nicefrac1\delta} \mr d w\, w^{d-1-\alpha_T+2\alpha_S} + o\left(\abs{\vl w}^{-\alpha_T + 2\alpha_S}\right) = \mc O\left(\delta^{\min(\alpha_T-d, 2\alpha_S)}\right).\label{eq:scalingsecondterm2}
\end{multline}

Finally, putting \eref{eq:scalingfirstterm}, \eref{eq:scalingsecondterm1} and \eref{eq:scalingsecondterm2} together,
\begin{equation}
\mb E\,\mr{MSE} = \mc O \left(\delta^{\min(\alpha_T-d, 2\alpha_S)}\right).
\end{equation}

The proof is concluded by considering that $\delta = \mc O\left(n^{-\nicefrac1d}\right)$.

In the case of a Gaussian kernel $K(\vl x) \propto \exp(-\abs{\vl x}^2/(2\sigma^2))$ --- and therefore $\tl K(\vl w) \propto \exp(-\sigma^2\abs{\vl w}^2/2)$ --- one has to redo the calculations starting from \eref{eq:mseprefinal}, but the final result can be easily recovered by taking the limit $\alpha\to+\infty$ (Gaussian kernels decay faster than any power law).

\end{proof}

\section{Proofs of lemmas}\label{app:prooflemmas}

\textbf{Lemma 1}
\emph{Let $K(\vl x, \vl x^\prime)$ be a translation-invariant isotropic kernel such that $\tl K(\vl w) = c \abs{\vl w}^{-\alpha} + o\left(\abs{\vl w}^{-\alpha}\right)$ as $\abs{w}\to\infty$ and $\abs{\vl w}^d \tl K(\vl w)\to0$ as $\abs{w}\to0$. If $\alpha > d+n$ for some $n\in\mb Z^+$, then $K(\vl x) \in C^n$, that is, it is at least $n$-times differentiable.}

\begin{proof}
The kernel is rotational invariant in real space ($K(\vl x) = K(\abs{\vl x})$) and therefore also in the frequency domain. Then, calling $\hat\epsilon_1 = (1, 0, \dots)$ the unitary vector along the first dimension $x_1$,
\begin{equation}
K(x) \propto \int\mr{d}\vl w\, e^{i\vl w\cdot \hat\epsilon_1 x} \tl K(\abs{\vl w}).
\end{equation}
It follows that
\begin{multline}
\left\lvert\partial^m K(x)\right\rvert \propto \left\lvert\int\mr{d}\vl w\, (\vl w\cdot\hat\epsilon_1)^m e^{i\vl w\cdot \hat\epsilon_1 x} \tl K(\abs{\vl w})\right\rvert < \int\mr{d}\vl w\, |\vl w\cdot\hat\epsilon_1|^m |\tl K(\abs{\vl w})| \propto \\
\propto \int_0^\infty \mr{d}w\, w^{d-1+m} |\tl K(w)| \int_0^\pi\mr{d}\phi_1 |\cos(\phi_1)|^m \propto \int_0^\infty \mr{d}w\, w^{d-1+m} |\tl K(w)|.
\end{multline}
We want to claim that this quantity is finite if $m\leq n$. Convergence at infinity requires $m<\alpha-d$, that is always smaller than or equal to $n$ because of the hypothesis of the lemma. Convergence in zero requires that $w^{d+m}|\tl K(w)|\to0$, and we want this to hold for all $0\leq m<\alpha-d$, the most constraining one being the condition with $m=0$.
\end{proof}\vspace{1em}

\textbf{Lemma 2}
\emph{Let $Z\sim\mc N(0,K)$ be a $d$-dimensional Gaussian random field, with $K\in C^{2n}$ being a $2n$-times differentiable kernel. Then $Z$ is $n$-times differentiable in the sense that
\begin{itemize}
\item derivatives of $Z(\vl x)$ are a Gaussian random fields;
\item $\mb E \partial_{x_1}^{n_1}\cdots\partial_{x_d}^{n_d} Z(\vl x) = 0$;
\item $\mb E \partial_{x_1}^{n_1}\cdots\partial_{x_d}^{n_d} Z(\vl x) \cdot \partial_{x_1}^{n_1^\prime}\cdots\partial_{x_d}^{n_d^\prime} Z(\vl x^\prime) = \partial_{x_1}^{n_1+n_1^\prime}\cdots\partial_{x_d}^{n_d+n_d^\prime} K(\vl x - \vl x^\prime) < \infty$ if the derivatives of $K$ exist.
\end{itemize}
In particular, $\mb E \partial_{x_i}^{m} Z(\vl x) \cdot \partial_{x_i}^{m} Z(\vl x^\prime) = \partial_{x_i}^{2m}K(\vl x-\vl x^\prime) < \infty\ \forall m\leq n.$}

\begin{proof}
Derivatives of $Z(\vl x)$ are defined as limits of sums and differences of the field $Z$ evaluated at different points, therefore they are Gaussian random fields too, and furthermore it is straightforward to see that their expected value is always $0$ if the field itself is zero centered.

The correlation can be computed via induction. Assume that $\mb E \partial_{x_1}^{n_1}\cdots\partial_{x_d}^{n_d} Z(\vl x) \cdot \partial_{x_1}^{n_1^\prime}\cdots\partial_{x_d}^{n_d^\prime} Z(\vl x^\prime) = \partial_{x_1}^{n_1+n_1^\prime}\cdots\partial_{x_d}^{n_d+n_d^\prime} K(\vl x - \vl x^\prime)$ holds true. Then, if we increment $n_1$:
\begin{multline}
\mb E \partial_{x_1}^{n_1 + 1}\cdots\partial_{x_d}^{n_d} Z(\vl x) \cdot \partial_{x_1}^{n_1^\prime}\cdots\partial_{x_d}^{n_d^\prime} Z(\vl x^\prime) = \\
= \lim_{h\to0} h^{-1} \mb E\left[ \partial_{x_1}^{n_1}\cdots\partial_{x_d}^{n_d} Z(\vl x + h\hat\epsilon_1) - \partial_{x_1}^{n_1}\cdots\partial_{x_d}^{n_d} Z(\vl x) \right] \cdot \partial_{x_1}^{n_1^\prime}\cdots\partial_{x_d}^{n_d^\prime} Z(\vl x^\prime) = \\
= \lim_{h\to0} h^{-1} \left[ \partial_{x_1}^{n_1+n_1^\prime}\cdots\partial_{x_d}^{n_d+n_d^\prime} K(\vl x - \vl x^\prime + h\hat\epsilon_1) - \partial_{x_1}^{n_1+n_1^\prime}\cdots\partial_{x_d}^{n_d+n_d^\prime} K(\vl x - \vl x^\prime) \right] = \\
= \partial_{x_1}^{n_1+1+n_1^\prime}\cdots\partial_{x_d}^{n_d+n_d^\prime} K(\vl x - \vl x^\prime).
\end{multline}
Of course by symmetry the same can be said about the increase of any other exponent.
To conclude the induction proof we simply recall that by definition $\mb E Z(\vl x)Z(\vl x^\prime) = K(\vl x - \vl x^\prime)$.
\end{proof}

\section{RKHS hypothesis ans smoothness}\label{app:rkhssmooth}
It is important to note the high degree of smoothness underlying the RKHS hypothesis. Consider for instance realizations $Z(\vl x)$ of a Teacher Gaussian process with covariance $K_T$ and assume that they lie in the RKHS of the Student kernel $K_S$ (notice that they never belong to the RKHS of the same kernel $K_T$), namely
\begin{equation}
\mb E \abs{Z}_{K_S}^2 = \mb E \int \mr{d}\vl x\mr{d}\vl\, y Z(\vl x)K_S^{-1}(\vl x-\vl y)Z(\vl y) = \int\mr{d}\vl w\, \tl K_T(\vl w)\tl K_S^{-1}(\vl w)<\infty.
\end{equation}
If the Teacher and Student kernels decay in the frequency domain with exponents $\alpha_T$ and $\alpha_S$ respectively, convergence requires $\alpha_T>\alpha_S+d$, and $K_S(\vl 0)\propto\int\mr{d}\vl w\,\tl K_S(\vl w)<\infty$ (true for many commonly used kernels) implies $\alpha_S>d$. Then using \lref{lm:lemma1} and \lref{lm:lemma2} we can conclude that the realizations $Z(\vl x)$  must be at least $\lfloor\nicefrac{d}2\rfloor$-times mean-square differentiable to be RKHS.

\section{Asymptotic limit of the PDE approximation}\label{app:approxresult}

In this appendix we show how to recover the prediction of \tref{th:theorem} using the approach presented in~\cite{bordelon2020spectrum}, that can be applied to a generic target function (not necessarily Gaussian nor evaluated only on a regular lattice). They derive their main formula, that we write below, by computing the amount of generalization error due to each eigenmode of the (Student) kernel. In order to carry out the calculations they introduce a partial differential equation that they solve with two different approximations.

In the following we denote by $\lambda_1\geq\cdots\geq\lambda_\rho\geq\cdots$ the eigenvalues of the kernel, and by $\phi_\rho(\vl x)$ the corresponding eigenfunctions. In~\cite{bordelon2020spectrum} they show that the generalization error can be written as:
\begin{align}
    &\mb E \mr{MSE} = \sum_\rho E_\rho(n)\label{eq:spectrumEg},\\
    &E_\rho(n) = \sum_\rho \frac{\mb E w_\rho^2}{\lambda_\rho} \left(\frac1{\lambda_\rho} + \frac{n}{t(n)}\right)^{-2} \left(1 - \frac{n \gamma(n)}{t(n)^2}\right)^{-1},\\
    &\gamma(n) = \sum_\rho \frac{\lambda_\rho^2}{\left(1 + \lambda_\rho\frac{n}{t(n)}\right)^2},\\
    &t(n) = \sum_\rho \frac{\lambda_\rho}{1 + \lambda_\rho \frac{n}{t(n)}}.
\end{align}
The term $\mb E w_\rho^2$ is the variance of the coefficients of the target function in the kernel eigenbasis, defined as:
\begin{equation}
    w_\rho = \lambda_\rho^{-\nicefrac12}\LA Z, \phi_\rho \RA,
\end{equation}
(the factor in front of the scalar product is to keep our notation consistent with that of \cite{bordelon2020spectrum}), to help the reader compare the two works. Notice that the variance can be computed with respect to an ensemble of target functions, but this ensemble may contain one deterministic function only.

In order to compute sums over the eigenmodes we will always replace them with integrals over eigenvalues. To do so, we must also introduce a density of eigenvalues $\mc D(\lambda)$: $\sum_\rho f(\lambda_\rho) \to \int\mr d\lambda\,\mc D(\lambda) f(\lambda)$. The asymptotic behavior of this density for small eigenvalues can be derived as follows (for a given kernel whose Fourier transform decays with an exponent $\alpha$):
\begin{equation}
    \mc D(\lambda) = \sum_\rho \delta(\lambda-\lambda_\rho) \sim \int\mr d^d\vl w\, \delta(\lambda - \abs{\vl w}^{-\alpha}) \sim \int_0^\infty \mr dw\, w^{d-1} \delta(\lambda-w^{-\alpha}) = \lambda^{-\theta}, \label{eq:densityspectrum}
\end{equation}
where we have defined the exponent $\theta \equiv 1 + \frac{d}\alpha$. Notice that $1<\theta<2$, and that of course this exponent depends on the kernel. We can use this density also to derive a scaling behavior of small eigenvalues: indeed, the $\rho$-th ($\gg1$) eigenvalue can be estimated by
\begin{equation}
    \rho \sim \int_{\lambda_\rho}^{\lambda_1} \mr d\lambda\, \mc D(\lambda) \sim \int_{\lambda_\rho}^{\lambda_1} \mr d\lambda \lambda^{-\theta} \sim \lambda_\rho^{-(\theta-1)}. \label{eq:rhotheigenvalue}
\end{equation}
The last equation follows from the fact that $\lambda_\rho \ll \lambda_1$ and that $\theta>1$.

We now have to estimate the asymptotic behavior of the implicitly defined function $t(n)$. It is easy to see that this function must go to $0$ as $n\to\infty$, therefore we can assume it small. Splitting the integral according to whether the denominator in the definition of $t(n)$ is dominated by the first or second term,
\begin{equation}
    t(n) \sim \int\mr d\lambda\, \lambda^{-\theta} \frac{\lambda}{1 + \lambda \frac{n}{t(n)}} \sim
    \int_0^{\frac{t(n)}{n}} \mr d\lambda\, \lambda^{-\theta} + \int_{\frac{t(n)}{n}}^{\lambda_1} \mr d\lambda\, \lambda^{-\theta} \frac{t(n)}{n} \sim \left(\frac{t(n)}{n}\right)^{2-\theta}.
\end{equation}
Therefore, $t(n) \sim n^{-\frac{2-\theta}{\theta-1}}$, and with a similar approximation we can also deduce that $\gamma(n) \sim n^{-\frac{3-\theta}{\theta-1}}$. Injecting all we know in the formula for the generalization error and splitting the integral we find
\begin{equation}
\label{222}
    \mb E \mr{MSE} \sim \sum_\rho \frac{\mb E w_\rho^2}{\lambda_\rho} \left(\frac1{\lambda_\rho} + n^\frac1{\theta-1}\right)^{-2}(1-n^0)^{-1} \sim
    \sum_\rho \frac{\mb E w_\rho^2}{\lambda_\rho} \left(\frac1{\lambda_\rho} + \frac1{\lambda_n}\right)^{-2} \sim
    \lambda_n^2 \sum_{\rho \leq n} \frac{\mb E w_\rho^2}{\lambda_\rho} + \sum_{\rho > n} \mb E w_\rho^2 \lambda_\rho.
\end{equation}
In the second equality we have used the fact that $\lambda_n\sim n^{-\frac1{\theta-1}}$ to introduce the $n$-th eigenvalue into the formula. Then, we approximated the sum by splitting it in two sums, one over the first $n$ eigenvalues ($\rho\leq n$, therefore $\lambda_\rho\geq\lambda_n$) and one over the remaining ones ($\rho>n$). Notice that the second sum is indeed the sum in \eref{eq:interpretmodes}.

Next we assume that $\mb E w_\rho^2$ behaves asymptotically as a power law with respect to small  eigenvalues, $\mb E w_\rho^2\sim \lambda_\rho^q$, with an exponent $q$ that can be either positive or negative. We can now compute each of the integrals in the previous equation:
\begin{align}
     \lambda_n^2 \sum_{\rho \leq n} \frac{\mb E w_\rho^2}{\lambda_\rho} &\sim \lambda_n^2 \int_{\lambda_n}^{\lambda_1} \mr d\lambda\, \lambda^{-\theta} \lambda^{q-1},\label{eq:firstterm}\\
    \sum_{\rho > n} \mb E w_\rho^2 \lambda_\rho &\sim \int_0^{\lambda_n} \mr d\lambda\, \lambda^{-\theta} \lambda^{q+1} \sim \lambda_n^{q-\theta+2} \sim n^{-\frac{q-\theta+2}{\theta-1}}.\label{eq:secondterm}
\end{align}
For the second integral to converge we have assumed that the exponent $q$ is larger than $\theta-1$. The first integral behaves differently according to whether $q>\theta$ or not: if $q>\theta$, the integral scales as $\lambda_n^2 \sim n^{-\frac2{\theta-1}}$; if $q<\theta$, then it scales as $\lambda_n^{2-\theta+q} \sim n^{-\frac{q-\theta+2}{\theta-1}}$. Therefore,
\begin{equation}
    \mb E\,\mr{MSE} \sim n^{-\frac{\min(q-\theta,0) + 2}{\theta-1}}.
\end{equation}

A consequence of \eref{eq:firstterm} and \eref{eq:secondterm} is that if $q<\theta$ (which always occur if the student is smooth enough, so that $\alpha$ characterizing the decay of the Fourier coefficient is small and $\theta$ is large), then the scaling of the generalization error is given by \eref{eq:secondterm} alone, and we recover \eref{eq:interpretmodes} from \eref{222}, justifying why this equation applies to real, non-Gaussian data.

Notice that if the target function is generated by a Teacher Gaussian process, the exponent $q$ takes the value $\frac{\theta-\theta_T}{\theta_T-1}$, where $\theta_T = 1+\frac{d}{\alpha_T}$ and $\alpha_T$ is the exponent characterising the decay of the Fourier transform of the Teacher kernel. With some manipulations we then recover our \tref{th:theorem}
\begin{equation}
    \mb E \mr{MSE} \sim n^{-\frac1d\min(\alpha_T-d,2\alpha_S)}.
\end{equation}

\section{Convergence of the spectrum of the Gram matrix}\label{app:otherplots}


\begin{figure}[b!]
\centering
\includegraphics[width=\textwidth]{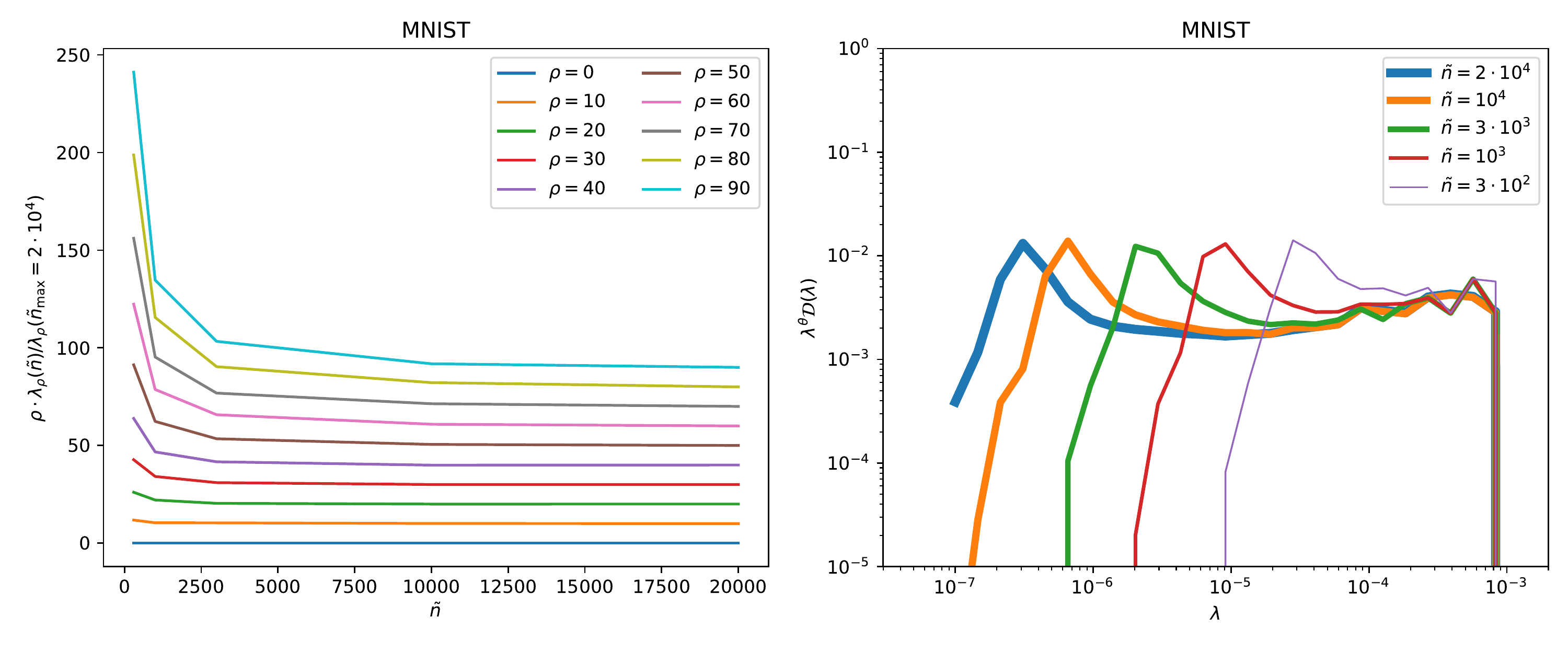}\vspace{-1em}
\includegraphics[width=\textwidth]{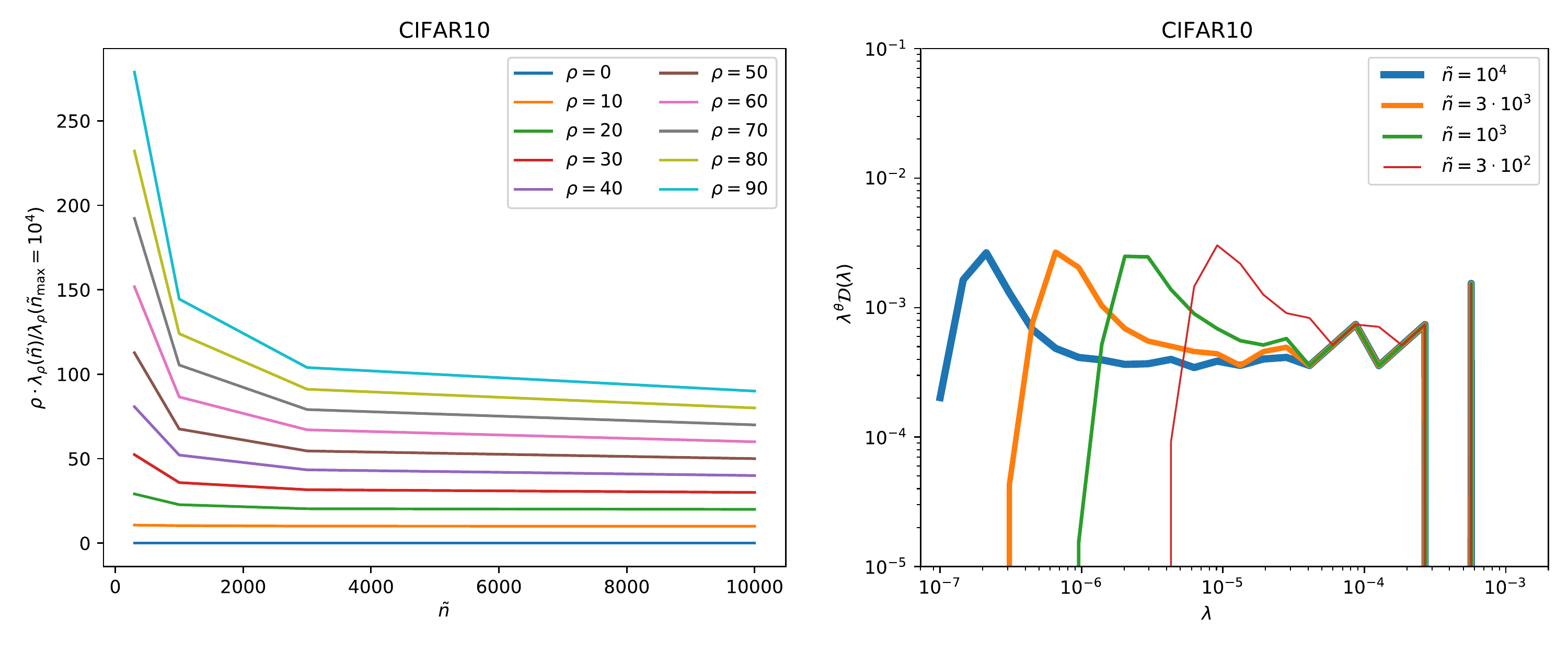}
\caption{\textbf{Left:} we plot the first eigenvalues $\lambda_\rho$ of Gram matrices of size $\tl{n}$, rescaled by the corresponding eigenvalue of the largest Gram matrix (top row is MNIST, bottom row is CIFAR10). Increasing $\tl{n}$ the eigenvalues are expected to converge, and indeed these ratios asymptote to one. We are plotting one eigenvalue every 10 for the first 100 eigenvalues. In order to make the plot clearer we have multiplied each curve by a factor $\rho$, equal to the eigenvalue index. \textbf{Right:} Density of eigenvalues of the Gram matrix, for several sizes $\tl{n}$, for MNIST (top) and CIFAR10 (bottom). The density is divided by the predicted asymptotic behavior $\left(\lambda_\rho^S\right)^{-\theta}$, with $\theta = 1 + \frac{\alpha}{d_\mr{eff}}$. For a Laplace kernel $\alpha_S = d_\mr{eff}+1$ and for the effective dimension we used the values extracted in \sref{sec:randomandrealdatascaling}, resulting in $\theta\approx1.937$ for MNIST and $\theta\approx1.972$ for CIFAR10. This plot shows that the density of eigenvalues converges when $\tl{n}$ increases, and that the predicted power law is consistent with observations.\label{fig:distlambda}}

\end{figure}

\end{document}